\documentclass[sn-nature]{sn-jnl}

\usepackage{tcolorbox} 
\usepackage{graphicx}%
\usepackage{multirow}%
\usepackage{amsmath,amssymb,amsfonts}%
\usepackage{amsthm}%
\usepackage{mathrsfs}%
\usepackage[title]{appendix}%
\usepackage{xcolor}%
\usepackage{textcomp}%
\usepackage{manyfoot}%
\usepackage{booktabs}%
\usepackage{algorithm}%
\usepackage{algorithmicx}%
\usepackage{algpseudocode}%
\usepackage{listings}%
\usepackage{cleveref}
\usepackage{subcaption}
\usepackage{pifont}
\usepackage{soul}

\newcommand{\xmark}{\ding{55}}
\tcbuselibrary{skins} 

\usepackage{makecell}

\theoremstyle{thmstyleone}%
\theoremstyle{thmstyletwo}%

\theoremstyle{thmstylethree}%
\newcommand{\eg}{e.\,g.}
\newcommand{\ie}{i.\,e.}

\newcommand{\cf}{{cf.\,}}

\begin{document}

\title[Affective Computing Has Changed: The Foundation Model Disruption]{Affective Computing Has Changed: The Foundation Model Disruption}


%
%

\author*[1,2,3,4,5]{\fnm{Bj\"orn}\sur{Schuller}}\email{schuller@ieee.org}
\author[1,3]{\fnm{Adria}\sur{Mallol-Ragolta}}
\author[5,6]{\fnm{Alejandro}\sur{Peña Almansa}}
\author[1,3,5]{\fnm{Iosif}\sur{Tsangko}}
\author[1,5,7]{\fnm{Mostafa M.}\sur{Amin}}
\author[1]{\fnm{Anastasia}\sur{Semertzidou}}
\author[1]{\fnm{Lukas}\sur{Christ}}
\author[1]{\fnm{Shahin}\sur{Amiriparian}}

%
%
%
\affil[1]{\small{CHI -- Chair of Health Informatics, MRI, Technical University of Munich, Germany}}
\affil[2]{\small{MDSI -- Munich Data Science Institute, Germany}}
\affil[3]{\small{MCML -- Munich Center for Machine Learning, Germany}}
\affil[4]{\small{GLAM -- Group on Language, Audio, \& Music, Imperial College London, UK}}
\affil[5]{\small{EIHW -- Chair of Embedded Intelligence for Health Care \& Wellbeing, University of Augsburg, Germany}}
\affil[6]{\small{School of Engineering, Universidad Autonoma de Madrid, Spain}}
\affil[7]{\small{AI R\&D Team, SyncPilot GmbH, Germany}}


\abstract{The 
dawn of Foundation Models has on the one hand revolutionised a wide range of research problems, and, on the other hand, democratised
the access and use of AI-based tools by the general public. We even observe an incursion of these models into disciplines related to human psychology, such as the Affective Computing domain, suggesting their affective, emerging capabilities. In this work, we aim to raise awareness of the power of Foundation Models in the field of Affective Computing by synthetically generating and analysing multimodal affective data, focusing on vision, linguistics, and speech (acoustics). 
We also discuss some 
fundamental problems, such as 
ethical issues and regulatory aspects, related to the use of Foundation Models in this research area.}




\maketitle

\section{Introduction}
\label{sec:introduction}

%
``\textit{The world of Affective Computing has changed. I see it in the vision modality. I read it in the linguistic modality. I hear it in the speech modality. Much that once was is outdated...}'' 
This quote, which might sound slightly familiar to the J.\,R.\,R.\ Tolkien's readers, 
aims to literary exemplify how the disruption of Foundation Models (FM) might be 
impacting the Affective Computing research as we knew it. Before centring the discussion on this topic, we summarise where we come from. 

The Affective Computing research can be broadly clustered into three main categories: recognition of affect,
generation of affective content, and response to affect~\cite{picard1997affective}. The development of systems with affective 
features opens a vast arsenal of use cases, ranging from digital psychology recognising depression~\cite{deshpande2017depression} to security applications (\eg, stress prediction in driving~\cite{healey2005detecting}). However, Affective Computing applications are mostly centred in Human-Computer and Human-Robot Interaction (HCI/HRI)~\cite{spezialetti2020emotion, liu2017facial, crumpton2016survey}, where the ability to interpret the human affect
is not only desirable to improve the communication~\cite{tan2022speech,amiriparian2023speech}, but a necessary capability to correctly understand the message. 
Early work in the field of psychology indicated that human affect is communicated in a multimodal manner through physical channels such as facial expressions, language, or voice
~\cite{mehrabian1968communication}. Consequently, the Affective Computing community has paid significant attention to visual~\cite{li2022deep, takalkar2018survey}, 
linguistic~\cite{birjali2021comprehensive, poria2019emotion}, and speech (acoustic)~\cite{schuller2018speech, khalil2019speech} data processing, including multimodal configurations~\cite{wang2022systematic, poria2017review}. 
Of the different dimensions of human affect, the emotional states have 
attracted major attention for its impact in a wide range of aspects of peoples' lives. Given such relevance, the Affective Computing community has been putting major efforts in the development of automated systems for the recognition and understanding of the human feelings. 
Early 
research in the field 
of emotion recognition 
relied on 
conventional Machine Learning (ML) pipelines, in which expert-crafted features 
were first extracted from the raw data such as pixels, words, or an audio signal, and then processed 
utilising traditional statistical methods; \eg, Support Vector Machines (SVM). 
The key to 
this conventional approach was 
to try to design 
a suitable 
set of features 
that captures 
emotional content; \ie, the hand-crafted features. 
In the visual domain, the emotional content was 
assumed to be mainly reflected in the facial expression information, from which the features were extracted; \eg, through Principal Component Analysis (PCA), or traditional Computer Vision (CV) techniques, such as Gabor wavelets or texture filters. 
In addition, prior knowledge of the Action Units (AUs) defined in the Facial Action Coding System (FACS)~\cite{ekman1978facial}  
helped in the design of these features. 
In linguistics, 
feature extraction typically relied on $n$-grams or bag-of-words 
-- a vector representation of text by sparse vectors that represent some form of vocabulary and some form of frequency of occurrence of the words in the current text --, 
after carefully preprocessing the raw text (\eg, by stemming or stopping).
In audio, several works 
noticed the rich emotional information reflected in the prosodic features~\cite{murray1993toward}, which in combination with spectral features 
-- \eg, the Mel Frequency Cepstral Coefficients (MFCC) --, formed the basis for Speech Emotion Recognition (SER). 

The success of Deep Learning (DL) in the early 2010's 
entailed a disruption to 
the entire field of Artificial Intelligence. 
The development of affective systems embraced the new trend as well, which was mainly marked by the use of 
representational learning via Deep Neural Networks (DNN). Models such as Convolutional Neural Networks (CNN) or Recurrent Neural Networks (RNN) proved to be extremely useful for extracting appropriate features when large amounts of data were available. 
Consequently, feature engineering took a backseat when data-driven approaches unleashed their potential.
This could be seen as the first disruption of the field: the learning of representations.
In the visual domain, end-to-end CNN-based systems accepted raw images as input, discarding the need for any feature engineering. 
In linguistics, 
complex preprocessing steps and statistical representations lost their relevance with the success of word embedding models, which could not only be utilised as input for a DNN model, but also contained semantic meaning in their own space, including affective information~\cite{caliskan2017semantics}. 
Moreover, some speech systems began to process 
the raw time signal \cite{Trigeorgis16-AFE} or the spectrogram representations 
with CNNs to take advantage of their enhanced representational capabilities.

Therefore, the efforts were 
ultimately shifted from experts crafting representations to experts choosing model architectures to learn these representations. A second, albeit less noted and exploited potential disruption came with the possibility of (neural) architecture search by reinforcement learning \cite{Zhang18-ELF} or more efficient approaches \cite{Rajapakshe24-EJOa}. This meant that in principle, once having affective (labelled) data, the representation could be learnt to then analyse 
affective data as well as the best architecture to do so. 

The data-driven disruption meant 
an improvement as well for synthesis of emotional data. In the visual domain, the introduction of adversarial learning~\cite{goodfellow2014gan} paved the way for Generative Adversarial Network- (GAN) based generative models, in which the emotional state was also controlled by explicit indicators~\cite{ding2018exprgan} or by style transfer~\cite{karras2020analyzing}. Some works improved the semantic control over the output by identifying and traversing emotional directions in the GAN latent space~\cite{balakrishnan2020towards}. 
For text, for instance, RNNs 
could similarly produce affective language \cite{ghosh2017affect}.
In speech (acoustics), where traditional approaches mostly relied on Gaussian Mixture Models (GMM) or Hidden Markov Models (HMM), 
new encoder-decoder architectures based on Long Short-Term Memory (LSTM) networks were devised, in which the emotional state could be controlled by both explicit labels~\cite{zhou2023speech} or style transfer from a reference~\cite{wang2018style}.

Yet, a critical issue in this era was the acquisition of such reliable, annotated data. The subjectivity of measuring inner emotion through self-assessment shifted the focus to perceived emotion. However, `measuring' outer perceived emotion usually requires several labellers to reduce uncertainty, hence coming at high effort and cost.
In addition, the lack of spontaneous data sources -- \eg, due to privacy restrictions -- favoured the use of acted/elicited, non-spontaneous data samples~\cite{lucey2010CK+, burkhardt2005database}. Whilst non- or less-spontaneous data eased the problem of data availability, it came with the drawback that the analysis of real-world emotion struggles with subtlety not met in training. Further, generated data samples may reflect unrealistic emotional responses if systems are only trained on such data.  
The acquisition of data from Internet sources (\eg, social media, films) allowed the collection of large ``in-the-wild'' databases~\cite{mollahosseini2019affectnet, kossaifi2021sewa}, whose annotations were obtained through semi-automatic methods, crowdsourcing, or based on the criteria of experts in affect. 

Nowadays, a third disruption is taking place in the Artificial Intelligence (AI) community. Whilst the first disruption established that (almost) no feature engineering was needed -- just a powerful model and annotated data --,  and the second allowed to learn also the optimal architectures, in current developments even 
specialised annotated 
affective 
data for training the models may no longer be needed, as 
affective computing abilities start to emerge in (general large data) pre-trained models as we will discuss in the following section. 
Nevertheless, curating high-quality sets of annotated data to some extent remains crucial to assess the performance of the models. 
New architectures, such as the Diffuser~\cite{rombach2022high} or the Transformer~\cite{vaswani2017attention}, together with self-supervised learning strategies~\cite{devlin2019bert} and inter-modality alignment techniques~\cite{radford2021learning}, have led to new FMs ~\cite{ouyang2022training, touvron2023llama, wang2023neural, podell2023sdxl}. These models have demonstrated surprising capabilities using prompt-based instructions, to the point that they can generate realistic data samples or perform zero-shot classification. 
The 
extent of the affective capabilities of these models, and the potential they open up, is still uncertain. Herein, we aim to shed some light on this topic, and explore how the emergence of FMs and the subsequent AI regulation are influencing the Affective Computing community.


\section{Emergence in Foundation Models} 
\label{sec:emergence}

One of the main characteristics of the Foundation Models (FMs) is that they are trained on a broad range of data, so that the resulting models can be utilised in a wide range of problems. 
In addition, they exploit large amounts of learning parameters. 
Given sufficient learning material, from a certain number of such parameters hence well trained, knowledge `emerges' in the FMs, and they achieve competitive performances in tasks they have not been specifically trained for. This can, however, be difficult to predict \cite{schaeffer2024emergent}. In this paper, we aim to investigate the `emergent' affective capabilities of FMs. Focusing on the vision (\cf \Cref{sec:image}), linguistics (\cf \Cref{sec:language}), and speech (acoustics) (\cf \Cref{sec:speech}) modalities, we assess the capabilities of current FMs to i) generate synthetic affective samples, from which we infer the conveyed emotions with pre-trained emotion recognition classifiers\footnote{Note that in principle, this can lead to a `closed loop', as we cannot be sure whether the data used in the pre-trained emotion classifiers has not also been used in the training of the Foundation Models, but generally, it is unlikely, as high-quality affective data are rarely freely available on the Internet due to their privacy restrictions.}, and ii) analyse well-established datasets in the field in a zero-shot manner. To favour the comparability among the different modalities investigated, we focus on the `Big Six' Ekman emotions~\cite{ekman1971constants} (\ie, fear, anger, happiness, sadness, disgust, and surprise), in addition to the neutral state.

\subsection{The Vision Modality Has Changed}
\label{sec:image}

\subsubsection{Generation}
\label{sec:image_generation}


%
In the visual domain, data synthesis started to obtain pseudorealistic results in the last decade thanks to the Generative Adversarial Network- (GAN) based models~\cite{wang2018style, ding2018exprgan}. 
Nowadays, a boost in the quality of the synthesised images has been achieved via text prompt inputs-based models, due to 
i) the CLIP model~\cite{radford2021learning} and ii) the Diffuser model~\cite{ho2020denoising}. The former was presented in conjunction with DALL-E~\cite{ramesh2021zero}, as a model to predict how well a given caption describes an image; \ie, a text-to-image alignment. 
The latter 
is a model that learns to reconstruct images by removing an added Gaussian noise through a Markov Chain. During inference, the model is able to generate new images from Gaussian noise, being more efficient than other generative architectures. Models such as Stable Diffusion (SD)~\cite{rombach2022high} or DALL-E 2~\cite{ramesh2022hierarchical} integrate the CLIP 
and the Diffuser 
models 
to efficiently synthesise images with high semantic control, an important feature to generate affective samples. Further, the decision to make generative models, 
such as 
SD, open-source has allowed the general 
public 
to discover the potential of the technology, 
ultimately speeding up its advancements.

\begin{figure}
    \centering
    \includegraphics[width = \textwidth]{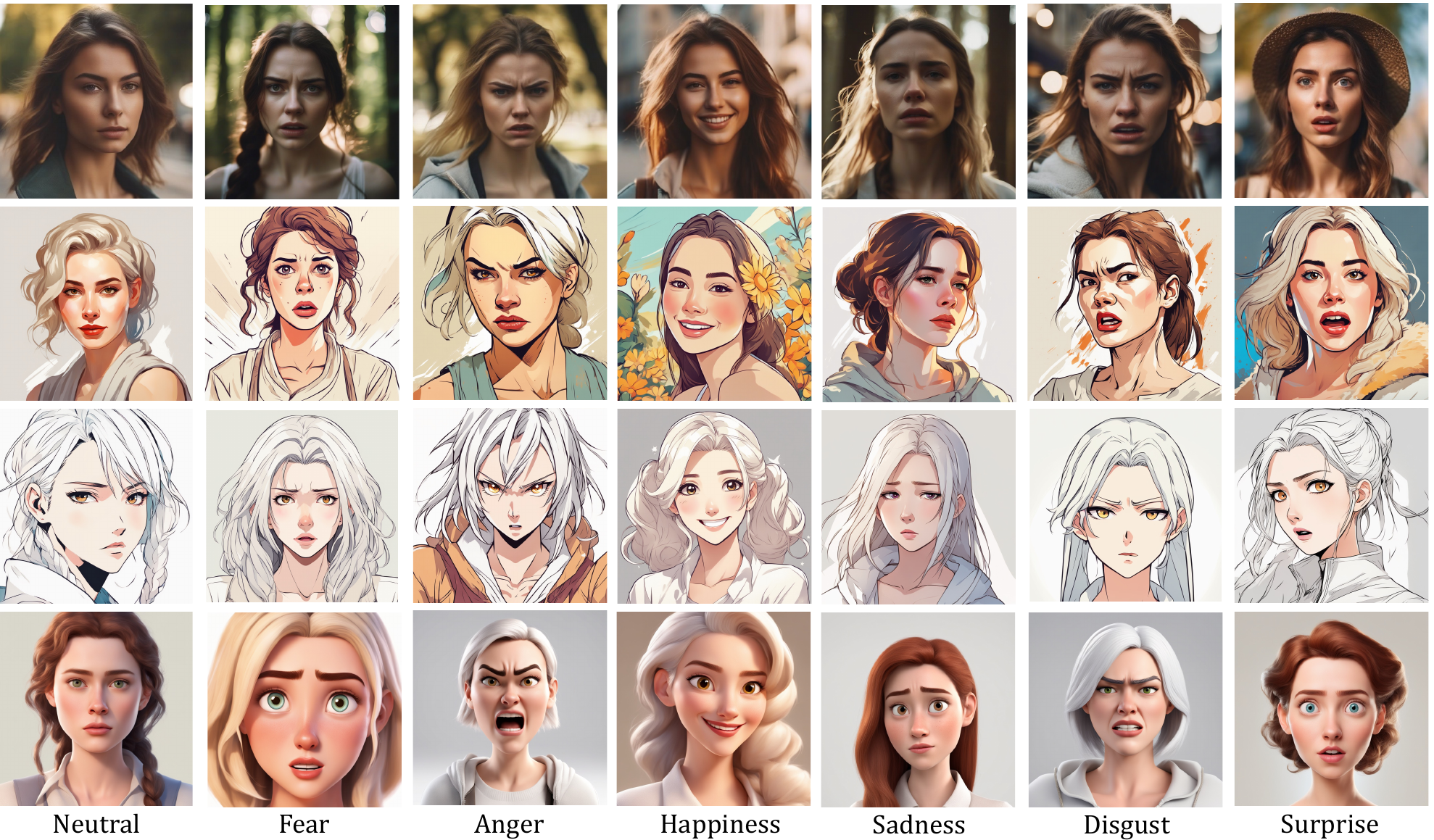}
    \caption{Synthetic facial images of a 
    white-skin, 
    young woman 
    conveying the `Big Six' Ekman emotions~\cite{ekman1971constants}, 
    in addition to the neutral state. All the images were generated with Stable Diffusion XL~\cite{podell2023sdxl}, conditioned on four different styles, namely photorealistic 
    (first row), cartoon-painting
    (second row), anime 
    (third row), and 3D (fourth row).}
    \label{fig:synthesized_images}
\end{figure}

We have leveraged one of the latest versions of SD~\footnote{\url{https://huggingface.co/stabilityai/stable-diffusion-xl-base-1.0}} -- \ie, Stable Diffusion XL (SDXL)~\cite{podell2023sdxl} -- 
to synthetically generate a face emotion dataset\footnote{For further information and access to the dataset, please contact the authors. }. This dataset is generated utilising prompts based on a fixed template with three sources of variation: i) the emotion, ii) the style (photorealistic, cartoon-painting, anime, and 3D), and iii) the demographic group. The template, along with the values explored for each attribute, are detailed in Table~\ref{tab:image_template}. 
Table~\ref{tab:synthetic_details} presents a summary of the gathered dataset. 
Although our emotion model is based on the `Big Six' Ekman emotions~\cite{ekman1971constants} plus the neutral state, we 
employed 
these 
basic 
emotions 
together with a higher intensity variation, as defined in Plutchik's model~\cite{plutchik2003emotions}, to further emphasise the desired affective states. Also, note that the generation process spans 18 different demographic groups; determined by age, biological sex, and skin tone. 
Visual examples for each emotion and style are provided in Figure~\ref{fig:synthesized_images} for the demographic group $<young, woman, white>$. For the case of the photorealistic style, we played with the background to generate some samples 
in 
realistic scenarios (\eg, outdoors, office, park). The generation process involved two experts, the SDXL base model and the SDXL refiner\footnote{\url{https://huggingface.co/stabilityai/stable-diffusion-xl-refiner-1.0/}}, with a total of 40 steps (80\,\% in the base model, 20\,\% in the refiner) and a guidance value of 7.5. Together with the desired prompt, we utilised a negative prompt to highlight what we do not want to see in the output. Once the images were generated, we filtered them according to four principles: 
the presence of disfigurations or artifacts, nudity, the quality of the emotion generated, and the plausibility of the style. For the sake of this experiment, only one annotator conducted 
this data curation process. 
%
%

\begin{table}[]
    \centering
    \caption{
    Attributes defined in the input prompts to synthesise emotional facial images with Stable Diffusion XL~\cite{podell2023sdxl}. 
    The prompt template 
    considers 
    three different sources of variation: the emotion, the style, and the demographic group. The latter is determined by three different demographic attributes: age, biological sex, and skin tone. We have also utilised a negative prompt, which includes all the styles that are not desired in the current synthesis.}
    \begin{tabular}{ll}
    \toprule
    \textbf{Attribute}&\textbf{Values}\\
    \midrule
    \makecell[l]{Prompt \\template} &\makecell[l]{Face image of a $<age>\,<sex>$ with $<skin>$ skin, with a $<emotion>$ face,\\ in a $<style>$ style, realistic eyes, white background, ultra quality, frontal picture,\\ looking at camera} \\[0.3cm]
    \makecell[l]{Negative\\ prompt}& disfigured, unrealistic eyes, blurry, b\&w, $<style>$\\[0.3cm]
    Emotion & \makecell[l]{\textit{neutral, fear and terror, anger and rage, happiness and joy, } \\ \textit{sadness and grief, disgust and loathing, surprise and amazement}}\\[0.3cm]
    Age & \textit{young, middle-aged, old} \\[0.25cm]
    Sex & \textit{man, woman}\\[0.25cm]
    Skin tone & \textit{white, brown, black}\\[0.25cm]
    Style &\textit{photorealistic, cartoon and painting, anime, 3D Pixar animation}\\[0.25cm]
    \bottomrule
    \end{tabular}
    \label{tab:image_template}
\end{table}

\begin{table*}[t]
\caption{Summary of the face images generated with the Stable Diffusion XL model~\cite{podell2023sdxl}. The prompts for the generation are detailed in Table~\ref{tab:image_template}.}
\label{tab:synthetic_details}
\begin{center}
    \begin{tabular}{lrrrr}
        \toprule
        \textbf{Emotion} & \textbf{Photorealistic} & \textbf{Cartoon} & \textbf{Anime} & \textbf{3D} \\[2pt]
        \midrule
        Neutral     &  $233$    & $132$ &  $131$ &  $143$ \\
        Fear        &  $185$    & $55$  &  $94$  &  $136$ \\
        Anger       &  $183$    & $165$ &  $119$ &  $119$ \\
        Happiness   &  $223$    & $202$ &  $118$ &  $142$ \\
        Sadness     &  $179$    & $156$ &  $137$ &  $99$  \\
        Disgust     &  $173$    & $90$  &  $56$  &  $40$  \\
        Surprise    &  $184$    & $147$ &  $120$ &  $139$ \\
        \midrule
        $\boldsymbol{\Sigma}$ & $1\,360$ & $947$ & $775$ & $818$ \\
        \bottomrule                
    \end{tabular}
\end{center}
\end{table*}

We 
now 
aim to automatically verify the affective quality of the generated facial images 
with Face Emotion Recognition (FER) models. We employ the manually annotated subset of the AffectNet dataset~\cite{mollahosseini2019affectnet} to develop these models. 
The images belonging to this dataset are annotated in terms of eleven emotions. Nevertheless, 
we select the images corresponding to the emotions fear, anger, happiness, sadness, disgust, and surprise in addition to the neutral class. We process the selected images with OpenFace~\cite{Baltruvsaitis16-OAO} to extract features related to a subset of the Action Units (AU) defined in the Facial Action Coding System (FACS). Specifically, OpenFace extracts 35 features per facial image, indicating the presence (0 or 1) and the intensity (in a scale from 0 -- not present -- to 5 -- present with maximum intensity) of a subset of the AUs. We discard the images that OpenFace fails to process; for instance, due to the absence of a face in the image. 
\Cref{table:affectnet} summarises the resulting data in terms of the number of images per emotion in the training and the validation partitions.

\begin{table*}[t]
\caption{Summary of the face images selected from AffectNet~\cite{mollahosseini2019affectnet} in the training and the validation partitions. }
\label{table:affectnet}
\begin{center}
    \begin{tabular}{lrr}
        \toprule
        \multicolumn{1}{c}{\textbf{Emotions}} & \multicolumn{1}{c}{\textbf{Training}} & \multicolumn{1}{c}{\textbf{Validation}} \\[2pt]
        \midrule
        Neutral     &  74\,873    & 500 \\
        Fear        &   6\,378    & 500 \\
        Anger       &  24\,881    & 500 \\
        Happiness   & 134\,411    & 500 \\
        Sadness     &  25\,458    & 500 \\
        Disgust     &   3\,803    & 500 \\
        Surprise    &  14\,090    & 500 \\
        \midrule
        $\boldsymbol{\Sigma}$ & 283\,894 & 3\,500 \\
        \bottomrule                
    \end{tabular}
\end{center}
\end{table*}

We start our preliminary investigation by training FER models with Support Vector Classifiers (SVC), as these are considered a standard machine learning technique with excellent results in a wide range of problems. 
We compare their performance when utilising a linear and a Radial Basis Function (RBF) kernel. One challenge associated with the selected dataset is the imbalanced training samples in terms of the emotional classes (cf.\,\Cref{table:affectnet}), which impacts the performance of the trained models. To overcome this issue, 
we consider weighting the training data, so that the samples corresponding to the least represented classes have more importance than the samples corresponding to the most represented classes when training the models. We fine-tune our models optimising the regularisation parameter $C \in [10^{-2}, 10^{-1}, 1, 10, 10^{2}]$. The performance of the optimal models on the validation partition is depicted in \Cref{table:affectnet_results}.

We also contrast the performance of the SVC-based FER models with a state-of-the-art Vision Transformer~\cite{Dosovitskiy-AII} for Facial Expression Recognition\footnote{\url{https://huggingface.co/trpakov/vit-face-expression}} (ViT -- FER), trained on the Facial Expression Recognition 2013 (FER-2013) dataset~\cite{Goodfellow13-CIR}. In this case, we use the pre-trained model off-the-shelf -- without fine-tuning -- and evaluate its performance on the validation partition of our subset of AffectNet. The obtained results are reported in \Cref{table:affectnet_results}. 

\begin{table*}[t]
\caption{Performance summary of the trained Support Vector Classifier-based models for Face Emotion Recognition on the validation partition of the considered subset of AffectNet. We also include the performance of a state-of-the-art Vision Transformer for Facial Expression Recognition (ViT -- FER). We select the accuracy (ACC) as the metric to assess the model performances. }
\label{table:affectnet_results}
\begin{center}
    \begin{tabular}{llcrr}
        \toprule
        \multicolumn{4}{c}{\textbf{Model}} & \multicolumn{1}{c}{\textbf{ACC ($\%$)}}\\[2pt]
        \midrule
        & \multicolumn{1}{c}{\textit{Kernel}} & \textit{Weighted Samples} & \multicolumn{1}{c}{\textit{Optimal $C$}} & \\[2pt]
        \cmidrule{2-4}        
        \multirow{4}{*}{SVC} & Linear      & \xmark        &   10    & 27.3\\
        & Linear      & $\checkmark$  &  $10^{2}$ &  28.7\\
        & Rbf         & \xmark        & $10^{2}$ & 28.5 \\
        & Rbf         & $\checkmark$  & $10^{2}$ & 29.7 \\
        \midrule
        \multicolumn{4}{l}{ViT -- FER}          & 43.9 \\
        \midrule
        \multicolumn{4}{l}{Chance Level}        & 14.3 \\
        \bottomrule                
    \end{tabular}
\end{center}
\end{table*}

The best performance on the validation partition of the AffectNet dataset (\cf \Cref{table:affectnet_results}) is obtained with the ViT -- FER model. Thus, we use this model to assess the affective quality of the generated facial images. The results obtained from the conducted experiments exemplify the 
breakthrough 
of working with end-to-end approaches, operating on the raw images instead of on the features extracted from them. Nevertheless, it is also worth mentioning that AffectNet was collected in the wild, which may complicate the estimation of the AUs from the facial images and, in turn, worsen the performance of the SVC-based models. 

\begin{table*}[t]
\caption{Accuracy (ACC) and Unweighted Average Recall (UAR) scores obtained when analysing the facial images generated in the four different styles with the ViT -- FER pre-trained model. }
\label{table:generatedImages_results}
\begin{center}
    \begin{tabular}{lrrrr}
        \toprule
        & \multicolumn{4}{c}{\textbf{Generated Styles}} \\[2pt]
        \cmidrule{2-5}
        & \textbf{Photorealistic} & \textbf{Cartoon-Painting} & \textbf{Anime} & \textbf{3D}\\[2pt]
        \midrule
        ACC ($\%$) & 35.0 & 43.9 & 42.5 & 57.5 \\
        UAR ($\%$) & 30.9 & 38.3 & 39.1 & 49.5 \\
        \bottomrule                
    \end{tabular}
\end{center}
\end{table*}

\begin{figure}
\centering
\begin{subfigure}{0.24\textwidth}
    \includegraphics[width=\textwidth]{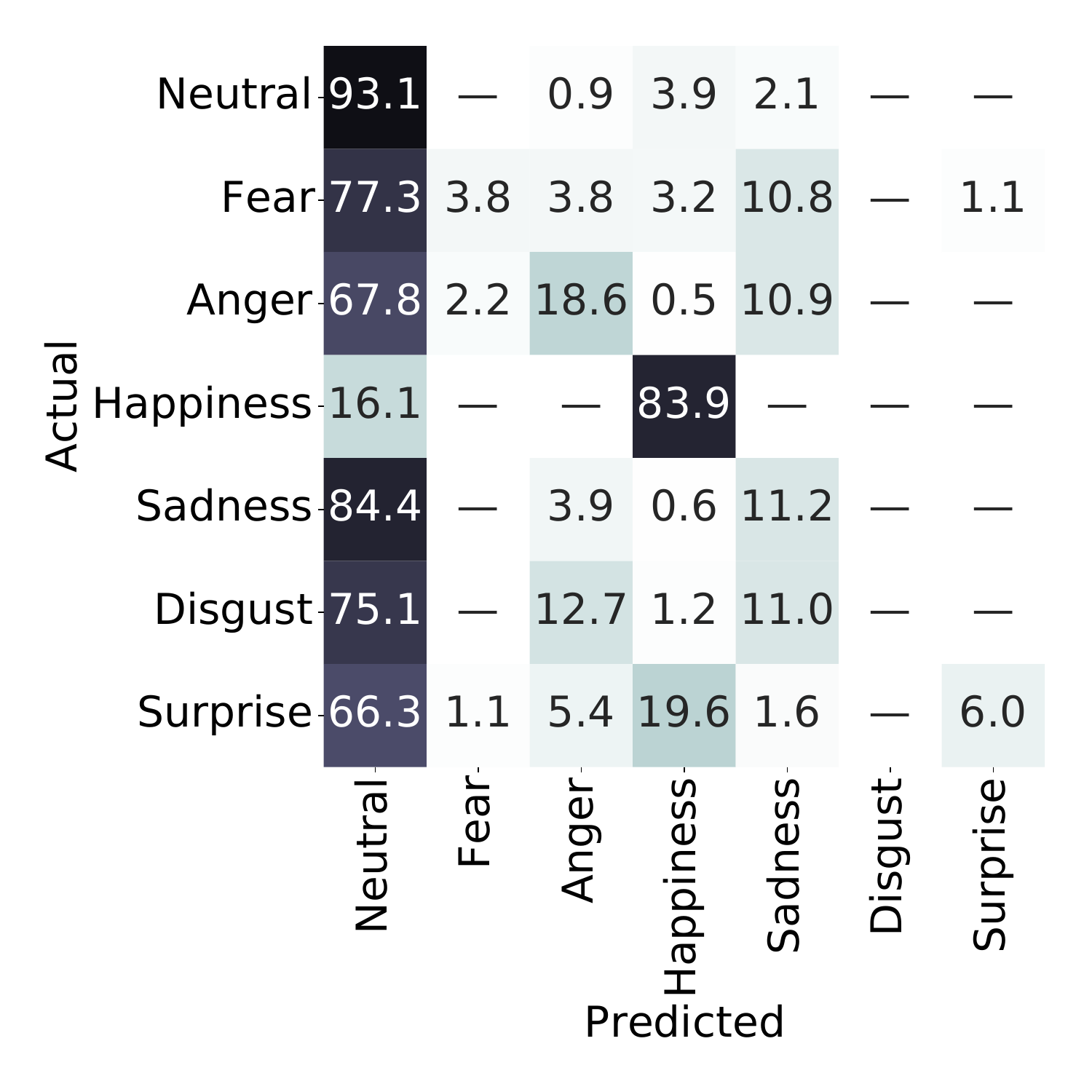}
    \caption{Photorealistic}
    \label{fig:generatedImg_CM_photorealistic}
\end{subfigure}
\hfill
\begin{subfigure}{0.24\textwidth}
    \includegraphics[width=\textwidth]{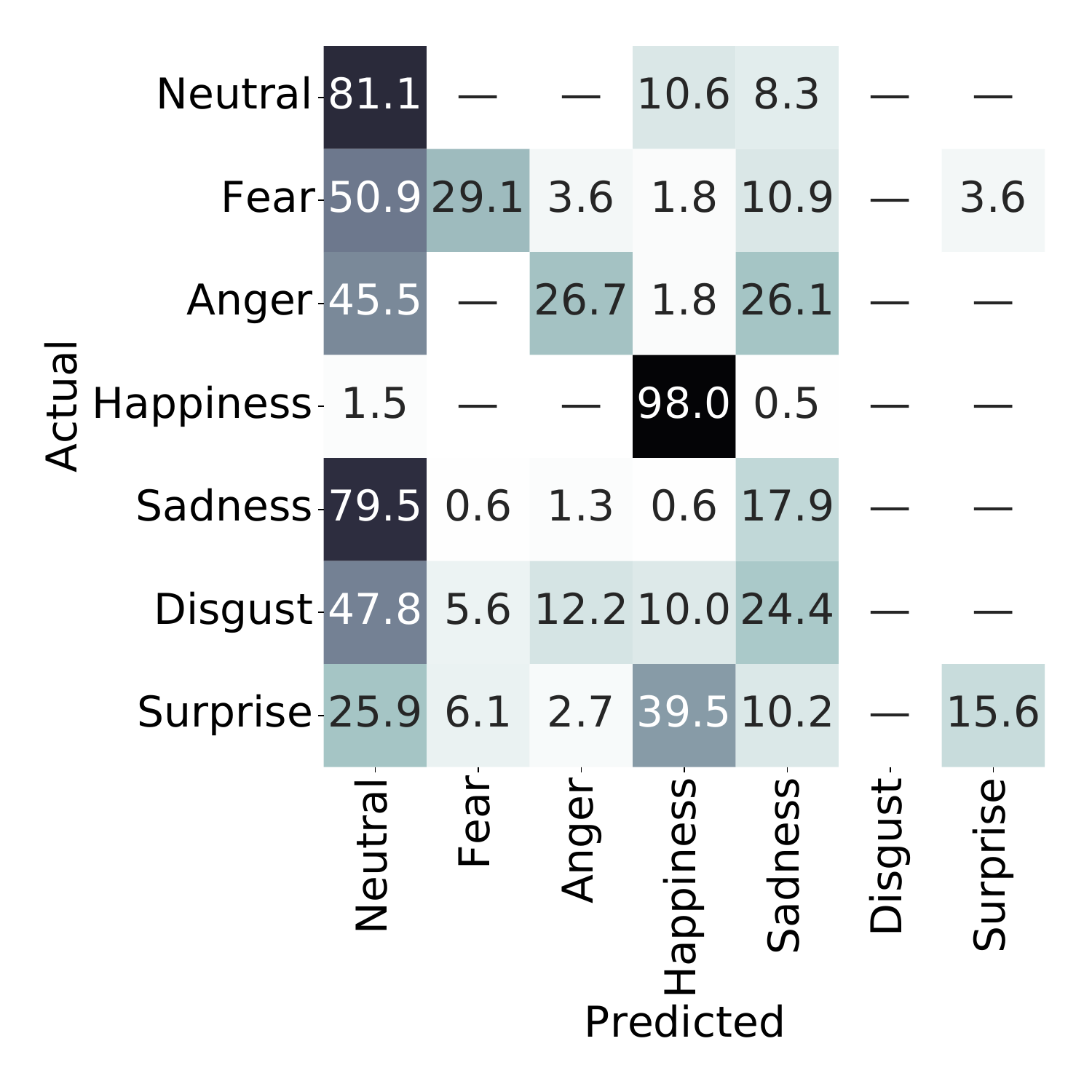}
    \caption{Cartoon-painting}
    \label{fig:generatedImg_CM_painting}
\end{subfigure}
\hfill
\begin{subfigure}{0.24\textwidth}
    \includegraphics[width=\textwidth]{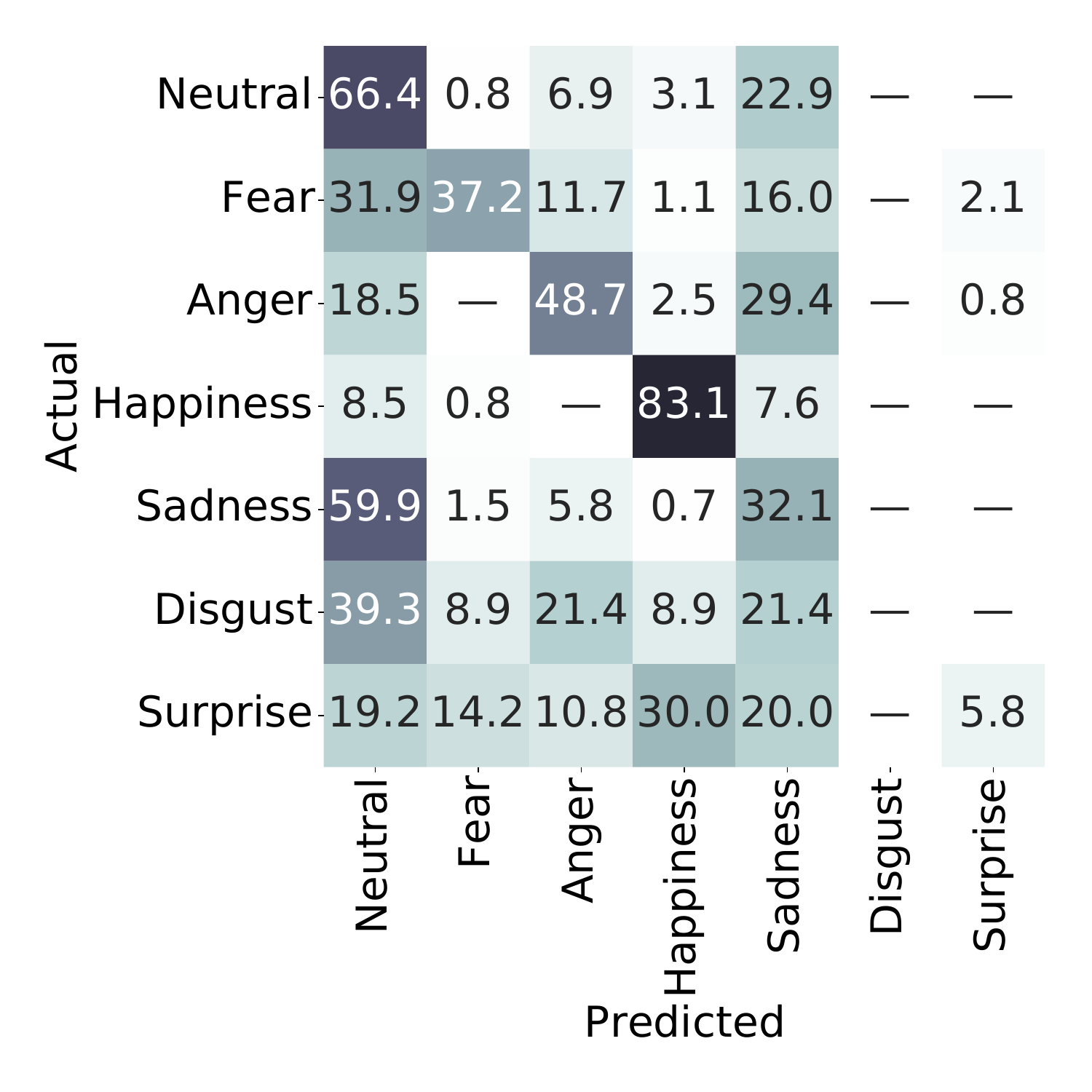}
    \caption{Anime}
    \label{fig:generatedImg_CM_anime}
\end{subfigure}
\hfill
\begin{subfigure}{0.24\textwidth}
    \includegraphics[width=\textwidth]{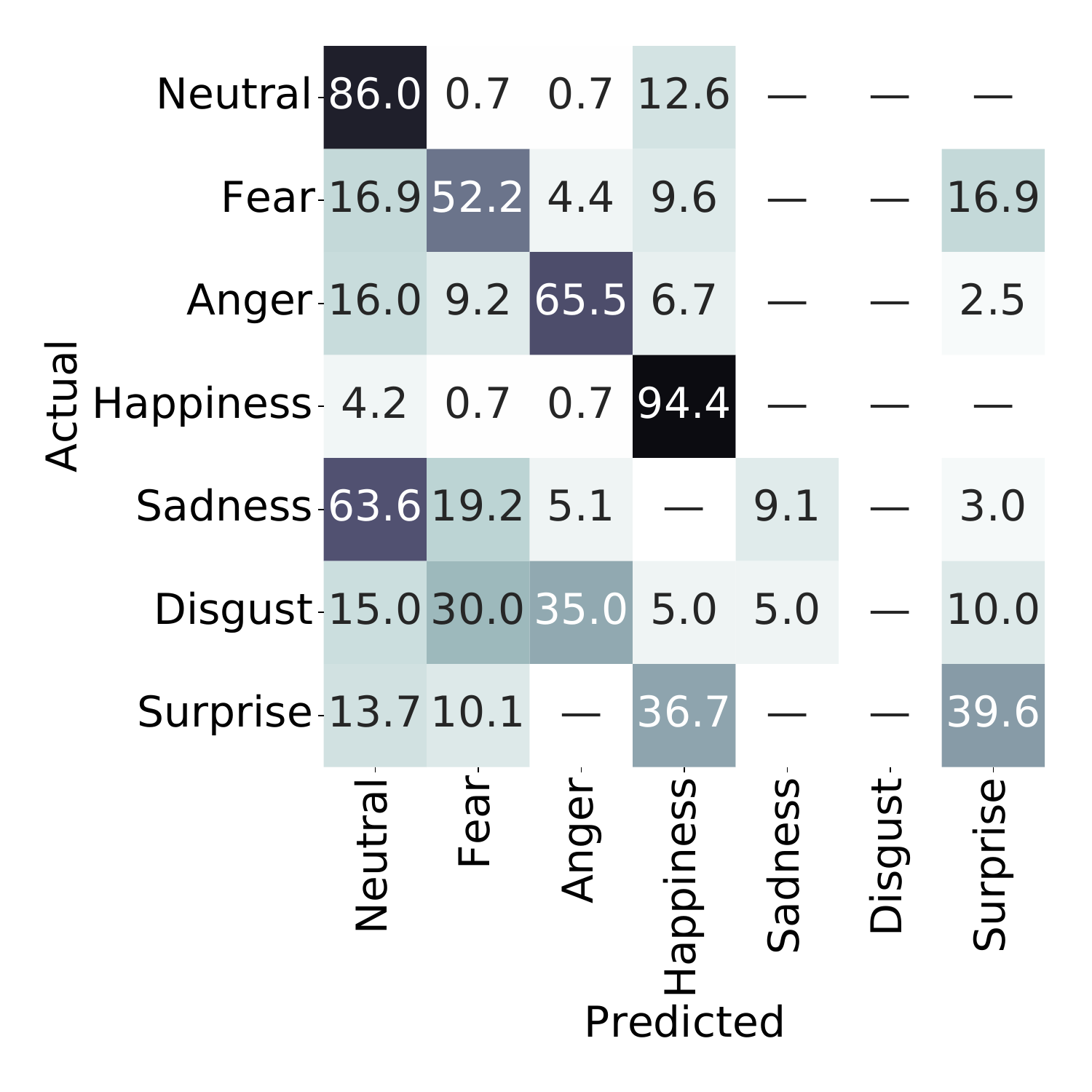}
    \caption{3D}
    \label{fig:generatedImg_CM_3d}
\end{subfigure}
\caption{Confusion matrices obtained by analysing the facial images generated according to the four different styles with the ViT -- FER pre-trained model.}
\label{fig:generatedImg_CMs}
\end{figure}

\Cref{table:generatedImages_results} summarises the results obtained when analysing the generated facial images with the four different styles utilising the ViT -- FER pre-trained model. \Cref{fig:generatedImg_CMs} presents the confusion matrices computed, comparing the model inferences and the ground truth. Across styles, the worst results are those of the photorealistic style, as denoted by both accuracy -- Weighted Average Recall (WAR) -- and Unweighted Average Recall (UAR)
\footnote{UAR is the sum of recall per class divided by the number of classes -- this reflects imbalances and is a standard measure in the field.}.
In contrast, the best results are obtained 
with 
the 3D style. 
Interestingly, in all the 4 styles both the neutral and the happiness 
emotions 
consistently obtain the best results. These classes are traditionally over-represented in face datasets (\eg, see the training set distribution of AffectNet in Table~\ref{table:affectnet}), probably due to the nature of the data sources~\cite{pena2021facial}. Consequently, it is expected for the generative model to be biased towards those 
emotions, which would also explain the difficulty to generate samples for classes like disgust or fear, 
which obtain the worst results. 

\subsubsection{Analysis}
\label{sec:image_analysis}

In order to evaluate the affective analytical capabilities of 
FMs 
in the image domain, 
we explore their performance in a zero-shot emotion 
recognition 
task 
under 
different configurations. Our experiments 
are conducted on the considered validation set of AffectNet~\cite{mollahosseini2019affectnet} (\cf \Cref{sec:image_generation}), as it is balanced, in-the-wild, and manually annotated. We compare 
three different approaches relying on model-prompting. In the first two, we start by extracting 
AU-based features 
with OpenFace~\cite{Baltruvsaitis16-OAO}, and 
we provide 
these 
features 
in a textual format as input 
to a FM. 
Recalling from \Cref{sec:image_generation}, OpenFace predicts both the presence and the intensity of a subset of AUs; hence, two approaches can be derived from its predictions. On the other hand, the third approach consists in 
directly feeding the images within the prompt of a FM. 
Note that the first two approaches can be addressed with a Language Foundation Model (LFM), for which we select the LLaMA2 7B model. 
We utilise a Multimodal Foundation Model (MFM) for the third scenario; specifically, the LLaVA1.5 7B model~\footnote{\url{https://huggingface.co/liuhaotian/llava-v1.5-7b}}~\cite{liu2023visual, liu2023improved}. 
This is an 
open-source MFM 
with visual capabilities trained by fine-tuning Vicuna\footnote{\url{https://lmsys.org/blog/2023-03-30-vicuna/}}, an already fine-tuned version of LLaMA, with GPT-4 generated data. 
The selected models allow them having the same 
number of parameters.

In \Cref{tab:image_analysis_results}, we present the results of the aforementioned 
scenarios. 
The prompts 
utilised for each scenario are detailed in \Cref{tab:image_analysis_prompt}. 
We also include 
in \Cref{tab:image_analysis_results} the results of the ViT -- FER model 
on the considered validation set of AffectNet for 
comparability 
purposes. Among the three zero-shot approaches, we obtain the best results when feeding the prompts with the raw images. 
Both AU-based prompt approaches exhibit accuracy results close to 
chance. 
The LLaVA model achieves an accuracy only 4 points 
below 
the ViT -- FER model, which was explicitly trained on the FER-2013 dataset to recognise 
emotions. 
This is an interesting result, which suggests that the LLaVA1.5 model presents emergent affective capabilities, despite not being specifically trained on affective computing tasks. 

\begin{table}[t!]
    \centering
    \caption{Prompts employed to perform zero-shot emotion 
    recognition 
    with Foundation Models in different scenarios. The prompts including (\ie, first two rows) AU information are injected in LLaMA2~\cite{touvron2023llama2}, while the prompt including the raw image is injected to 
    a LLaVA1.5 model~\cite{liu2023improved}.}
    \label{tab:image_analysis_prompt}
    \begin{tabular}{ll}
    \toprule
    \textbf{Approach}&\textbf{Prompt template}\\
    \midrule
    \makecell[l]{AU presence} &\makecell[l]{$<$s$>$[INST] $<<$SYS$>>$\\ You are a highly skilled Affective Computing system with an expertise in\\ accurately predicting emotion classes from Action Units. I will provide you a list\\ of Action Units present in a face. Your task is to answer the most likely emotion\\ class, without any further explanation. Please, provide only one of the following\\ classes as answer: Neutral, Fear, Anger, Happiness, Sadness, Disgust, Surprise. \\The question is, which is the most likely emotion if the following Action Units\\ are present$<</$SYS$>>$\\ 
“\{AU\}”.[/INST]\\
$\#\#\#$Response:} \\[2cm] 
    \makecell[l]{AU intensity} &\makecell[l]{$<$s$>$[INST] $<<$SYS$>>$\\You are a highly skilled Affective Computing system with an expertise in\\accurately predicting emotion classes from Action Units. I will provide you a \\JSON object with the intensities of the Action Units present in a face. Your\\ task is to answer the most likely emotion class, without any further explanation.\\Please, provide only one of the following classes as answer: Neutral, Fear,\\ Anger, Happiness, Sadness, Disgust, Surprise. \\The question is, which is the most likely emotion if the following Action Units\\ are present$<</$SYS$>>$\\ 
“\{AU\}”.[/INST]\\
$\#\#\#$Response:} \\[2cm]
    \makecell[l]{Image} &\makecell[l]{$<$image$>$\\USER: You are provided with a face image of a person. Classify the most likely \\emotional state depicted into one of the classes between brackets [Neutral,\\ Fear, Anger, Happiness, Sadness, Disgust, Surprise]\\
    ASSISTANT:} \\[0.3cm]

    \bottomrule
    \end{tabular}
\end{table}

\begin{table*}[t!]
\caption{Accuracy scores obtained with the LLaMA2~\cite{touvron2023llama2} and LLaVA1.5~\cite{liu2023improved} Foundation Models on the validation set of AffecNet~\cite{mollahosseini2019affectnet}. We have included as well the performance obtained with the ViT -- FER model~\cite{Dosovitskiy-AII} trained on FER-$2013$~\cite{Goodfellow13-CIR} for comparison purposes.}
\label{tab:image_analysis_results}
\begin{center}
    \begin{tabular}{llr}
        \toprule
        \textbf{Model}&\textbf{Input}&\textbf{ACC (\%)}\\
        \midrule
        LLaMA$2$ $7$B & Prompt with information of AU presence& $18.7$\\
        LLaMA$2$ $7$B & Prompt with description of AU presence& $13.4$\\
        LLaMA$2$ $7$B & Prompt with information of AU intensity& $17.9$\\
        LLaMA$2$ $7$B & Prompt with description of AU intensity& $16.8$\\
        LLaVA$1.5$ $7$B & Prompt with image & $39.3$\\
        LLaVA$1.5$ $7$B & Prompt with image and AU presence & $20.5$\\
        \midrule
        ViT -- FER & Image & $43.9$ \\
        \multicolumn{2}{l}{Chance level}& $14.3$\\
        \bottomrule                
    \end{tabular}
\end{center}
\end{table*}

\subsection{The Linguistic Modality Has Changed}
\label{sec:language}

\subsubsection{Generation}
\label{subsec:language-generation}

The evolution of text generation 
experienced 
an important shift in the mid-2010s 
with the development of neural models~\cite{mikolov2010recurrent}. 
The introduction of the Transformer model~\cite{vaswani2017attention} has revolutionised the Natural Language Processing (NLP) field and marked the beginning of the Large Language Models (LLM) era. Since then, this architecture has been established as the default backbone for 
the LLMs. In text generation, 
OpenAI's work with the Generative Pre-trained Transformer (GPT) models~\cite{radford2018improving,ouyang2022training}, culminating with the recent GPT-4~\cite{achiam2023gpt},  
has advanced text generation, paving the way for 
exhaustive 
research. Furthermore, developing open-source models, such as Meta's LLaMA models~\cite{touvron2023llama, touvron2023llama2}, 
democratised the 
access and 
fostered the 
innovation 
in the field. 
From an Affective Computing perspective, LLMs present a novel approach to inject and transfer emotional content  
in linguistic data, with recent works demonstrating their intrinsic emotional capabilities in a variety of domains~\cite{li2023large, broekens2023fine, amin2023will, amin2023wide, wang2023emotional}.

\begin{figure}
[t]\centering\vspace{-5em}\includegraphics[width=0.9\textwidth]{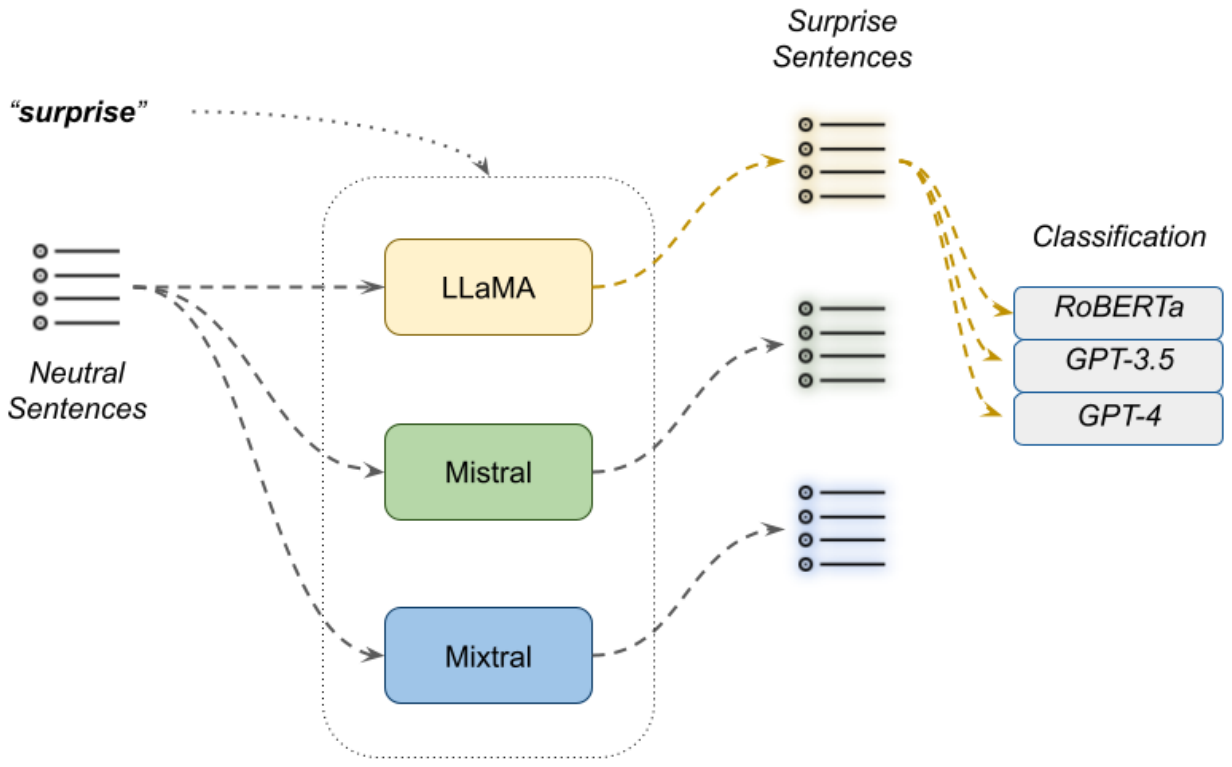}\vspace{-4em}\caption{Pipeline of the affective text style transfer process for generating the affective sentences with `surprise' as the prompted emotion. After that, we classify the synthesised sentences using RoBERTa, GPT-3.5, and GPT-4.} \label{fig:affective_text_style_transfer_pipeline} 
\end{figure}

In this section, we aim to investigate and leverage the affective style transfer capabilities of 
cutting-edge open-source LLMs. 
For this experiment, we select 
LLaMA2~\cite{touvron2023llama2} and Mistral~\cite{jiang2023mistral}, given their proven high performance in both language understanding and text generation tasks. Additionally, we include 
the Mixtral\footnote{\url{https://huggingface.co/mistralai/Mixtral-8x7B-v0.1}} LLM~\cite{jiang2024mixtral}, which utilises the Sparse Mixture of Experts (SMoE) method~\cite{fedus2022switch}. It is important to note that the implementation of SMoE in Mixtral notably influences its size. 
The LLaMA2 and Mistral LLMs each comprise 7 billion parameters.
We start with the text generation 
by compiling a corpus comprised of 122 human-curated neutral phrases.
The dataset encompasses various topics, ranging from mundane personal activities to formal professional interactions, thus providing a targeted platform to evaluate how well 
LLMs 
handle affective style transfer across diverse 
topics. 
Utilising the gathered corpus, 
we task the aforementioned LLMs (\ie, LLaMA2, Mistral, and Mixtral) to generate six emotional phrases from each original neutral phrase, conveying a specific target emotion: fear, anger, happiness, sadness, disgust, and surprise. 
The generation is described by the formula $\mathcal{M}(text, emotion), text \in \mathcal{N}, emotion \in \mathcal{E},$ where $\mathcal{M}$ is the LLM at hand, $\mathcal{N}$ the set of the neutral phrases, and $\mathcal{E} =\{fear, anger, happiness, sadness, disgust, surprise \}$ (the prompted emotion). We generate three 
synthetic sentences 
for each combination of text and emotion, yielding a 
corpus of $3*|\mathcal{N}|*|\mathcal{E}| = 2\,194$ emotional phrases\footnote{For further information and access to the dataset, please contact the authors. }. The data was synthesised with settings chosen to enhance variability and creativity.
The generation parameters used 
a temperature of 0.9, top-$p$ of 0.6, and a repetition penalty of 1.2.
An example of the generated 
sentences 
is shown in \Cref{tab:emotion_transformation_examples_surprise}, 
and \Cref{fig:affective_text_style_transfer_pipeline} visually summarises the followed pipeline. We observe that the models tend to exaggerate the injected emotion, adopting an over-dramatic style that results in more formal, yet affectively adapted phrases. Despite these exaggerated and dramatic adaptations, the primary objective was to investigate how these models perceive and express emotions under basic setup conditions. Our findings highlight the models' propensity to amplify emotional content, which was an anticipated aspect of this exploratory study.
\begin{table}[t]
    \centering
    \caption{Affective style transfer example towards the emotion `surprise' with the neutral phrase: \textit{``The weather is clear and sunny.''}}
    \label{tab:emotion_transformation_examples_surprise}
    \small  
    \begin{tabular}{p{2cm}p{9cm}}  
    \toprule
    \textbf{Model} & \textbf{Affective Phrase}\\
    \midrule
    Mixtral & \textit{Wow! What a surprise! The sky is astonishingly bright and clear today!} \\ [.5cm]
    \hline
    Mistral & \textit{The sudden emergence of unobstructed sunlight has taken me by complete astonishment!} \\[.5cm]
    \hline
    LLaMA & \textit{It comes as quite a shock to discover that the sky has transformed itself into such crystal clarity!} \\[.5cm]
    \bottomrule
    \end{tabular}
\end{table}

\begin{table}[t]
    \centering
    \caption{Statistics of the considered subset of the GoEmotions dataset. 
    }
    \begin{tabular}{lrrr}
    \toprule
         \textbf{Emotion}   & \textbf{Training} & \textbf{Validation} & \textbf{Test} \\
         \hline
         Neutral   &   12\,823  &   1\,592     & 1\,606 \\
         Fear      &    515   &   66       & 77   \\
         Anger     &    3\,878  &    485     &  520  \\
         Happiness &   12\,920  &   1\,668     &  1\,603 \\
         Sadness   &    2\,121  &   241      & 259   \\
         Disgust   &     498  &   61       & 76     \\
         Surprise  &    3\,553  &   435      & 449    \\
         \hline
         $\Sigma$  &   36\,308  &   4\,548     & 4\,590   \\
         \bottomrule
    \end{tabular}
    \label{tab:goemotions_summary_simplified}
\end{table}

In order to investigate the quality of the 
generated sentences 
by the LLMs, we implement two baseline models trained on the GoEmotions dataset~\cite{demszky2020goemotions}, 
a well-renowned corpus in the field and commonly utilised for benchmarking purposes due to its comprehensive labelling and categorisation of the emotions. 
The GoEmotions dataset consists of English Reddit comments annotated according to 27 distinct emotions, plus the neutral state, by 3 or 5 labellers each. 
%
%
Due to the nature of the annotations in the GoEmotions dataset, we 
begin by tailoring the data to meet the specific requirements of our experiments.
First, we select 
instances 
from the dataset 
annotated with a single emotion, in order to 
tackle the task 
as a single-label classification problem, instead of a multi-label classification one.
As previously, we adopt the `Big Six' Ekman emotions~\cite{ekman1971constants}, in addition to a seventh neutral state. 
This restructuring of the GoEmotions taxonomy to the Ekman taxonomy is achieved 
by aggregating the original labels into 
the targeted, 
broader categories~\cite{demszky2020goemotions}.
For example, emotions like annoyance and irritation, originally distinct, were grouped under `anger' to fit the Ekman model. 
\Cref{tab:goemotions_summary_simplified} summarises the statistics of the considered subset of the GoEmotions dataset. 

For the two baselines, we employ two different architectures: a Bi-directional Long Short-Term Memory (BiLSTM) network and a fine-tuned version 
of the RoBERTa-base model~\cite{liu2019roberta}.
Both models are trained on the 
selected subset of the 
GoEmotions dataset. The BiLSTM consists of two bidirectional LSTM layers with 128 units each.
It is trained with a learning rate of $5\times 10^{-3}$ and a batch size of 96 
for 40 epochs, while RoBERTa-base was fine-tuned at a conservative learning rate of $5\times 10^{-5}$ and a smaller batch size of 12. 
The models are trained on the training 
partition 
of the dataset, and the weights yielding the highest validation Unweighted Average Recall (UAR) are selected for each model.
The test scores for both baseline models (BiLSTM and RoBERTa) on the test 
partition 
of the GoEmotions dataset are shown in \Cref{tab:go-emotions-baselines}.
The results are consistent with~\cite{demszky2020goemotions}, but with some differences, given that we model the problem as a single-label classification, instead of the original multi-label classification task. 
Given the superior performance of RoBERTa, we utilise it for 
analysing the synthetically generated emotional sentences. 

\begin{table*}[!t]
    \centering
    \caption{Performance scores of the 
    implemented 
    models 
    when inferring the emotions (Ekman's `Big Six' in addition to the neutral state) conveyed by the sentences belonging to the test set of the GoEmotions dataset. 
    }
    \begin{tabular}{lrr}
    \hline
          \textbf{Model} & \textbf{ACC (\%)} & \textbf{UAR(\%)}\\
    \hline
BiLSTM  & 53.53 & 51.44 \\
RoBERTa & \textbf{69.22} & \textbf{62.82} \\
\hline
Chance Level & 14.29 & 14.29 \\
    \hline
    \end{tabular}
    \label{tab:go-emotions-baselines}
\end{table*}

\begin{figure}[!t]
\centering
\begin{subfigure}{0.32\textwidth}
    \includegraphics[width=\textwidth]{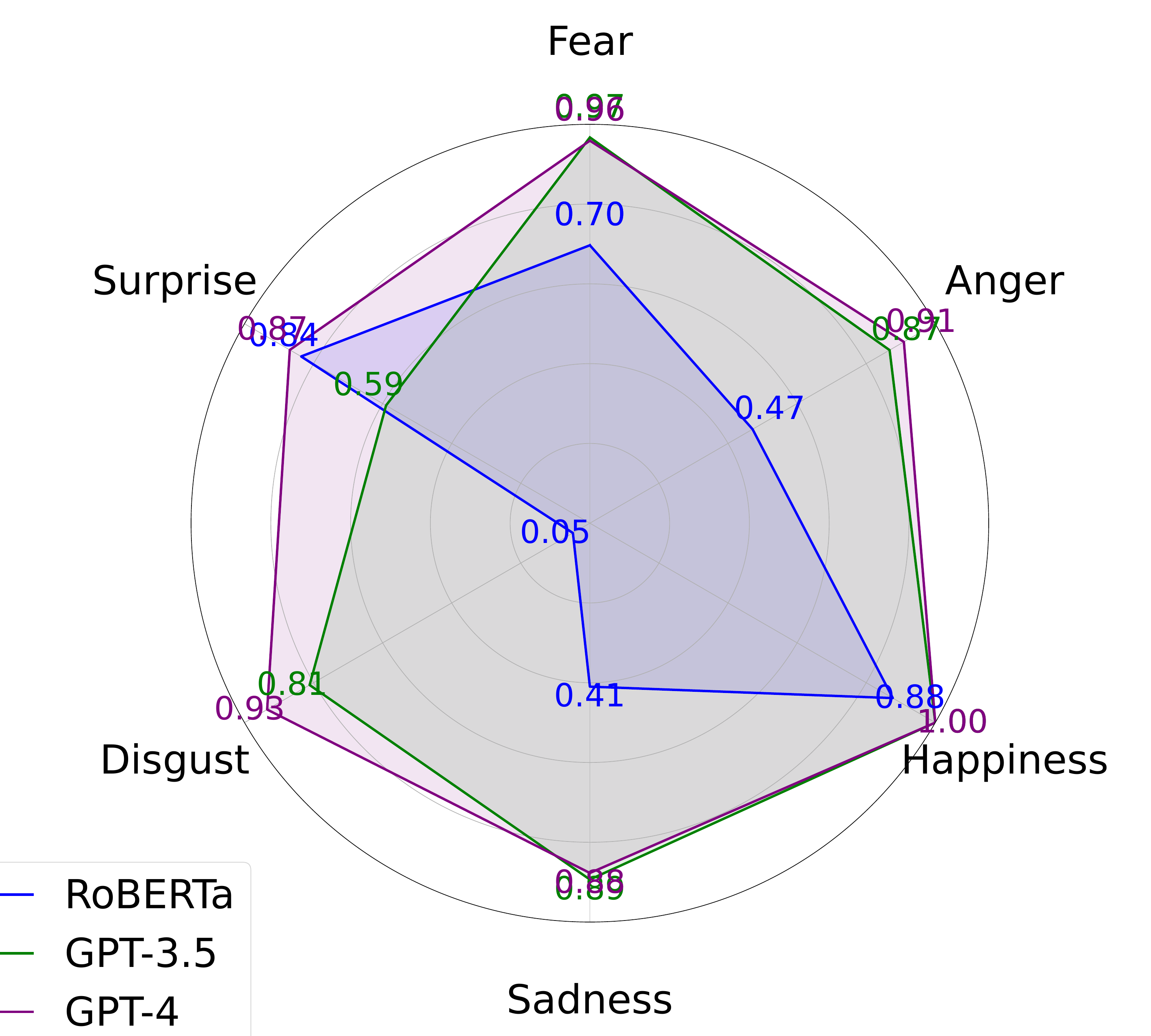}
    \caption{LLaMA2-based synthesis.}
    \label{fig:generatedSen_rd}
\end{subfigure}
\hfill
\begin{subfigure}{0.32\textwidth}
    \includegraphics[width=\textwidth]{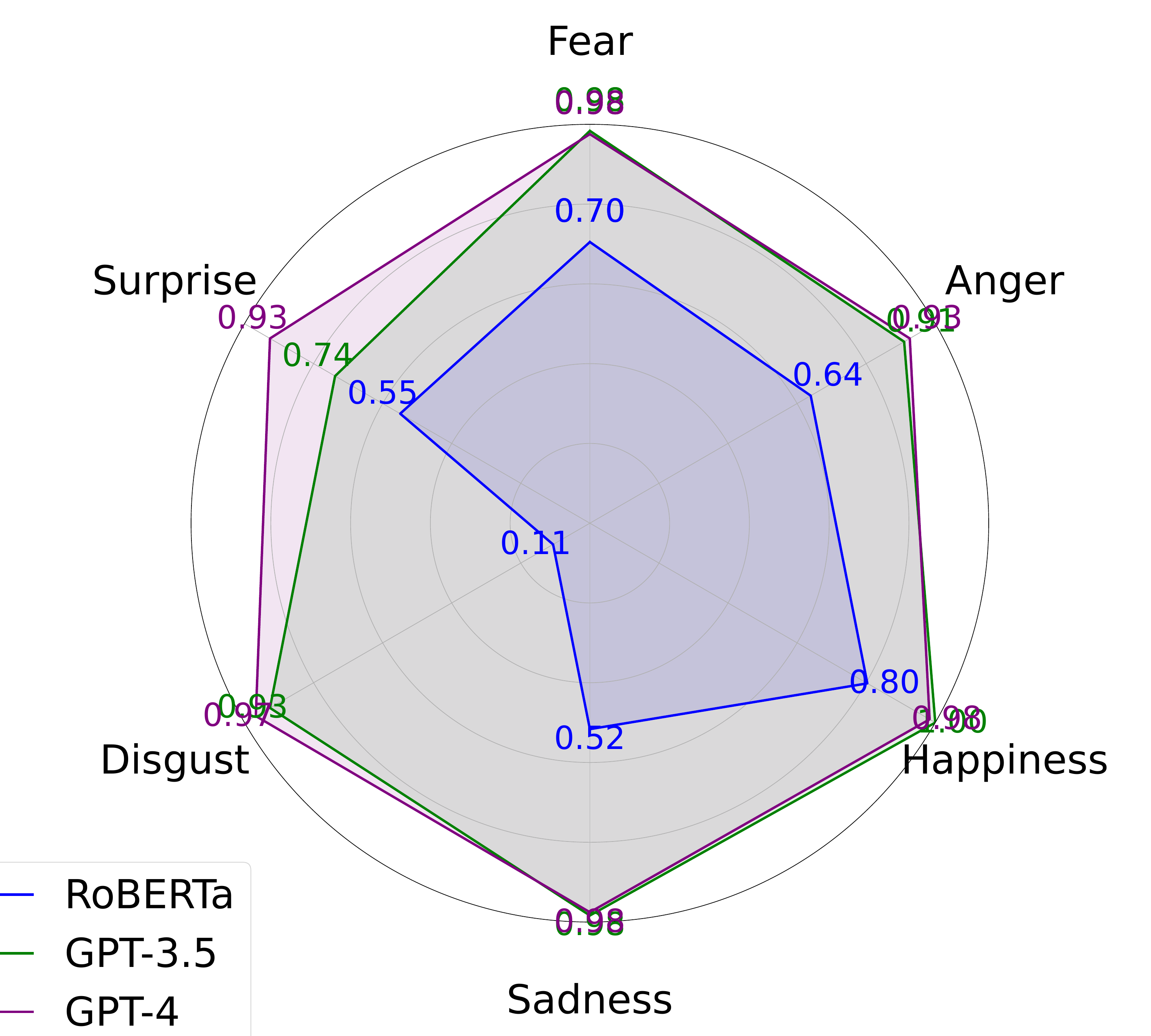}
    \caption{Mistral-based synthesis.}
    \label{fig:generatedSen_CM_mistral}
\end{subfigure}
\hfill
\begin{subfigure}{0.32\textwidth}
    \includegraphics[width=\textwidth]{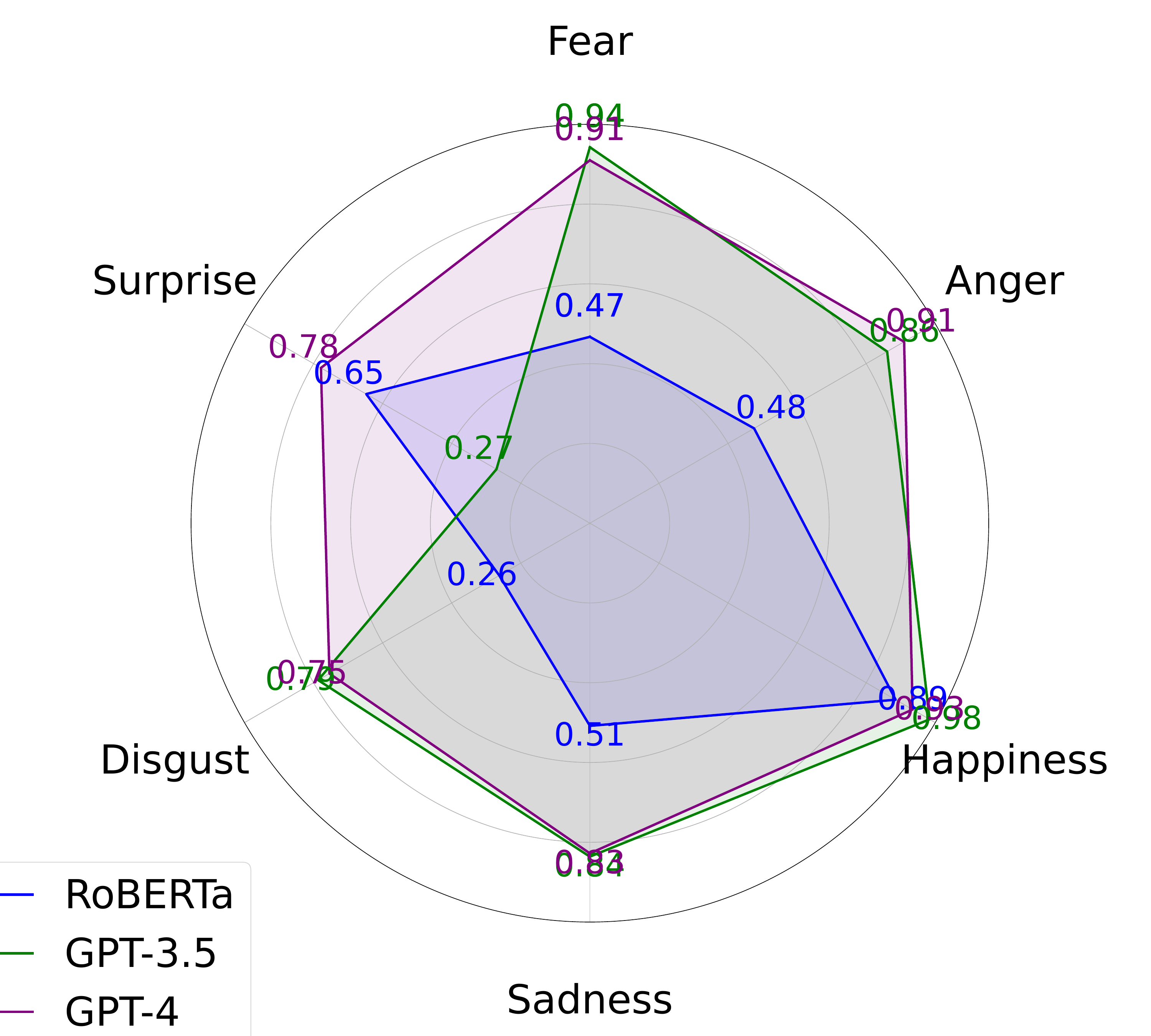}
    \caption{Mixtral-based synthesis.}
    \label{fig:generatedSen_third}
\end{subfigure}
\caption{
UAR scores obtained with the RoBERTa, GPT-3.5, and GPT-4 models when recognising the emotions conveyed by the synthetic sentences generated by LLaMA2 (left), Mistral (centre), and Mixtral (right).}
\label{fig:text-synthesis-radars}
\end{figure}

To evaluate the performance of the various LLMs on the emotion injection task, 
we test the generated sentences with the RoBERTa baseline model, 
in addition to GPT-4 as an approximation for human evaluation, and its weaker variant GPT-3.5. 
GPT-4 has 
shown superior performance in many affective computing problems \cite{amin2023wide}, often better than fine-tuned, specialised models, especially with problems 
related to 
sentiment or emotions.
\Cref{tab:gpt-prompts} demonstrates the prompt templates used for the GPT models, following a similar pattern like \cite{amin2023wide,amin2024prompt}.
The versions of GPT variants used are `gpt-3.5-turbo-0125'\footnote{https://platform.openai.com/docs/models/gpt-3-5-turbo} for GPT-3.5 and `gpt-4-turbo-2024-04-09'\footnote{https://platform.openai.com/docs/models/gpt-4-turbo-and-gpt-4} for GPT-4. 
The results of this evaluation are 
depicted 
in \Cref{fig:text-synthesis-radars}. Notably, these results can be considered to reflect 
a better agreement 
between models than with the ground truth labels, 
which are not human-annotated. 
However, the results of GPT-4 should be the closest to the human evaluations~\cite{zheng2024judging,amin2023wide}.

GPT-4 as the most superior model achieves very high UAR scores on all six emotions.
Its inferior model GPT-3.5 achieves slightly worse results in most cases, 
but it experiences a performance drop in the recognition of surprise. 
On the other hand, RoBERTa has a different behaviour in comparison.
It is generally much worse than 
GPT-4 and GPT-3.5 
in most of the cases, 
but 
it 
obtains 
a higher score than GPT-3.5 for the surprise emotion 
with the LLaMA2- and the Mixtral-generated sentences. 
Additionally, RoBERTa is showing very low performance for disgust, 
which seems to be a weakness of the model, consistent with the results on the GoEmotions dataset~\cite{demszky2020goemotions}. 

\begin{figure}[!t]
\centering
\begin{subfigure}{0.24\textwidth}
    \includegraphics[width=\textwidth]{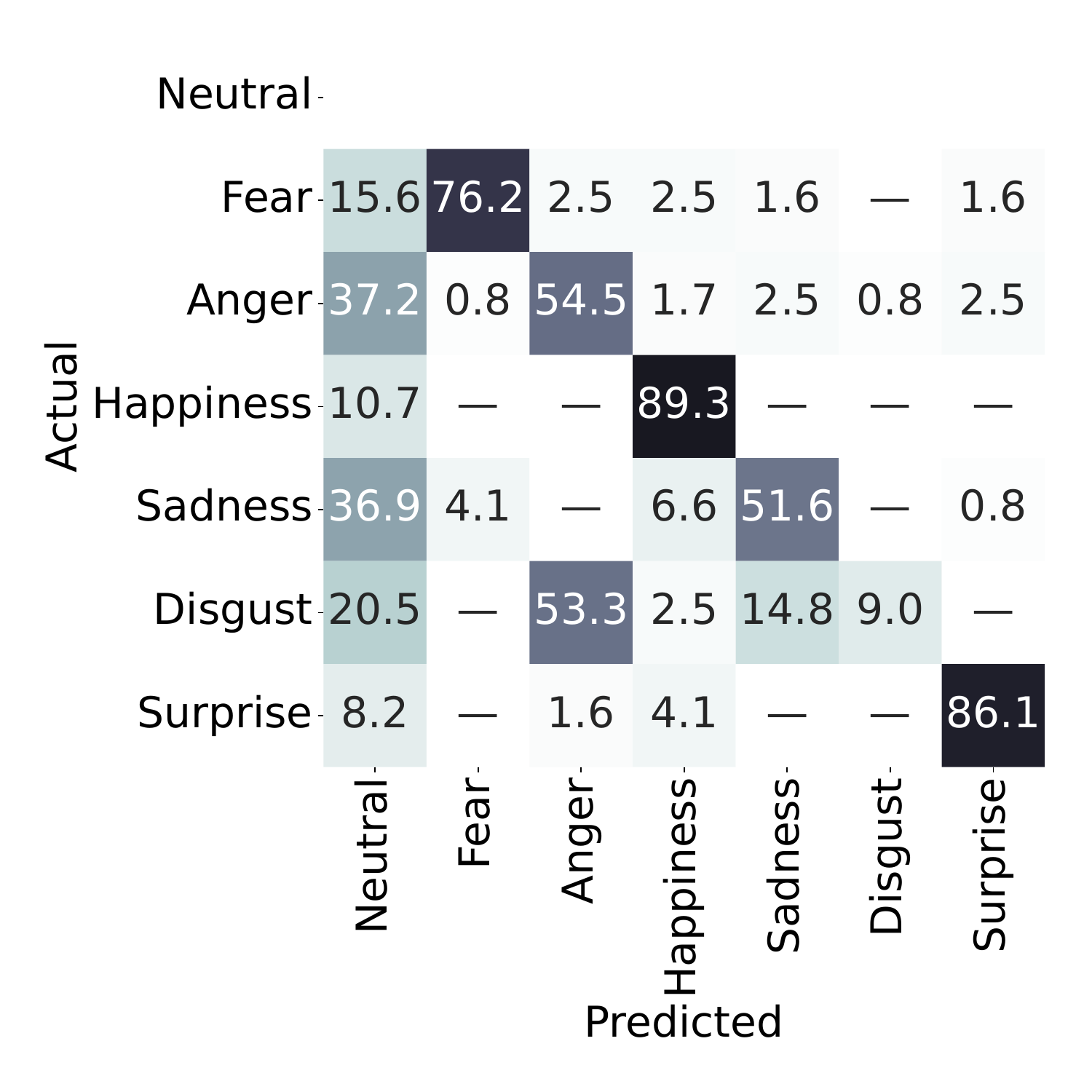}
    \caption{LLaMA2}
\end{subfigure}
\hfill
\begin{subfigure}{0.24\textwidth}
    \includegraphics[width=\textwidth]{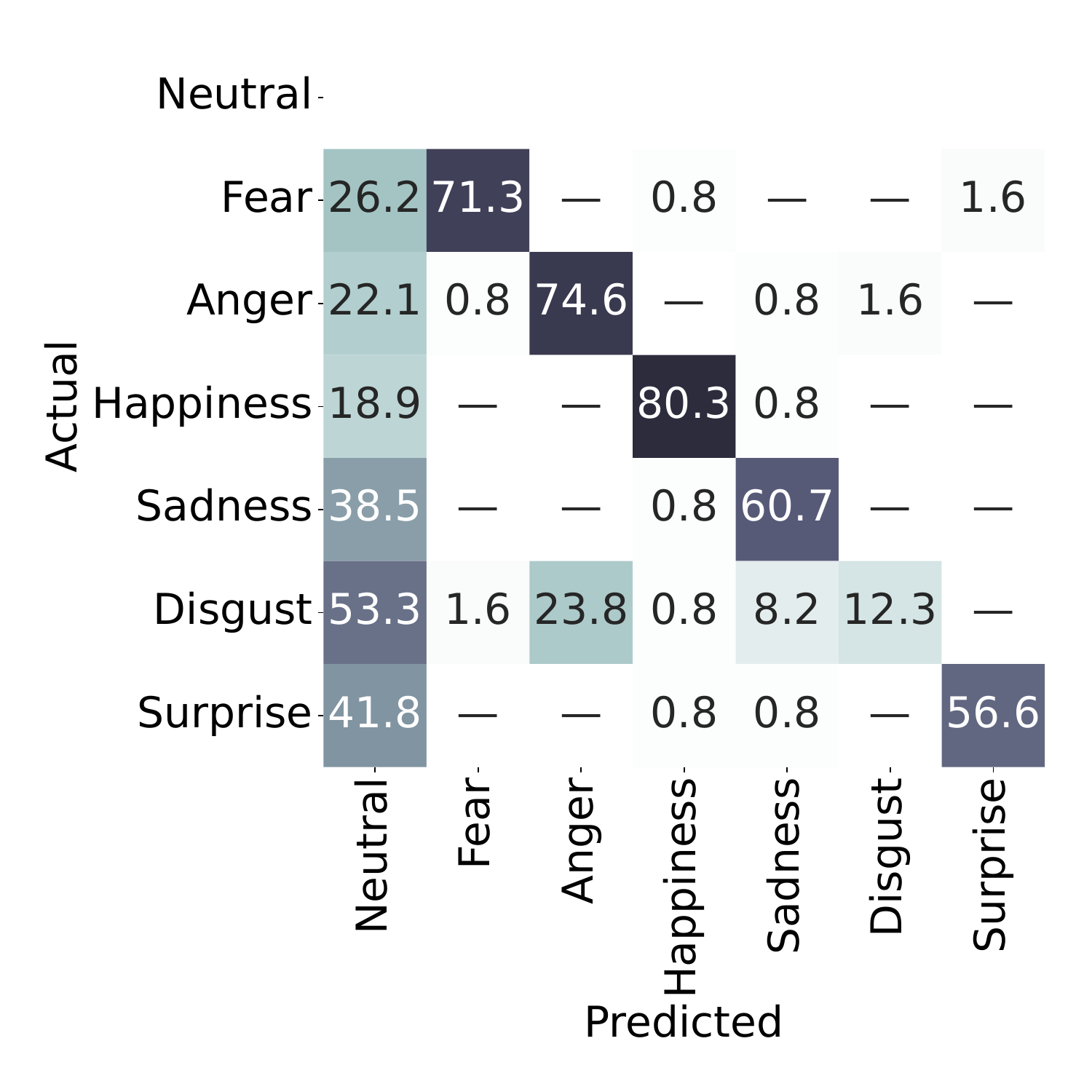}
    \caption{Mistral}
\end{subfigure}
\hfill
\begin{subfigure}{0.24\textwidth}
    \includegraphics[width=\textwidth]{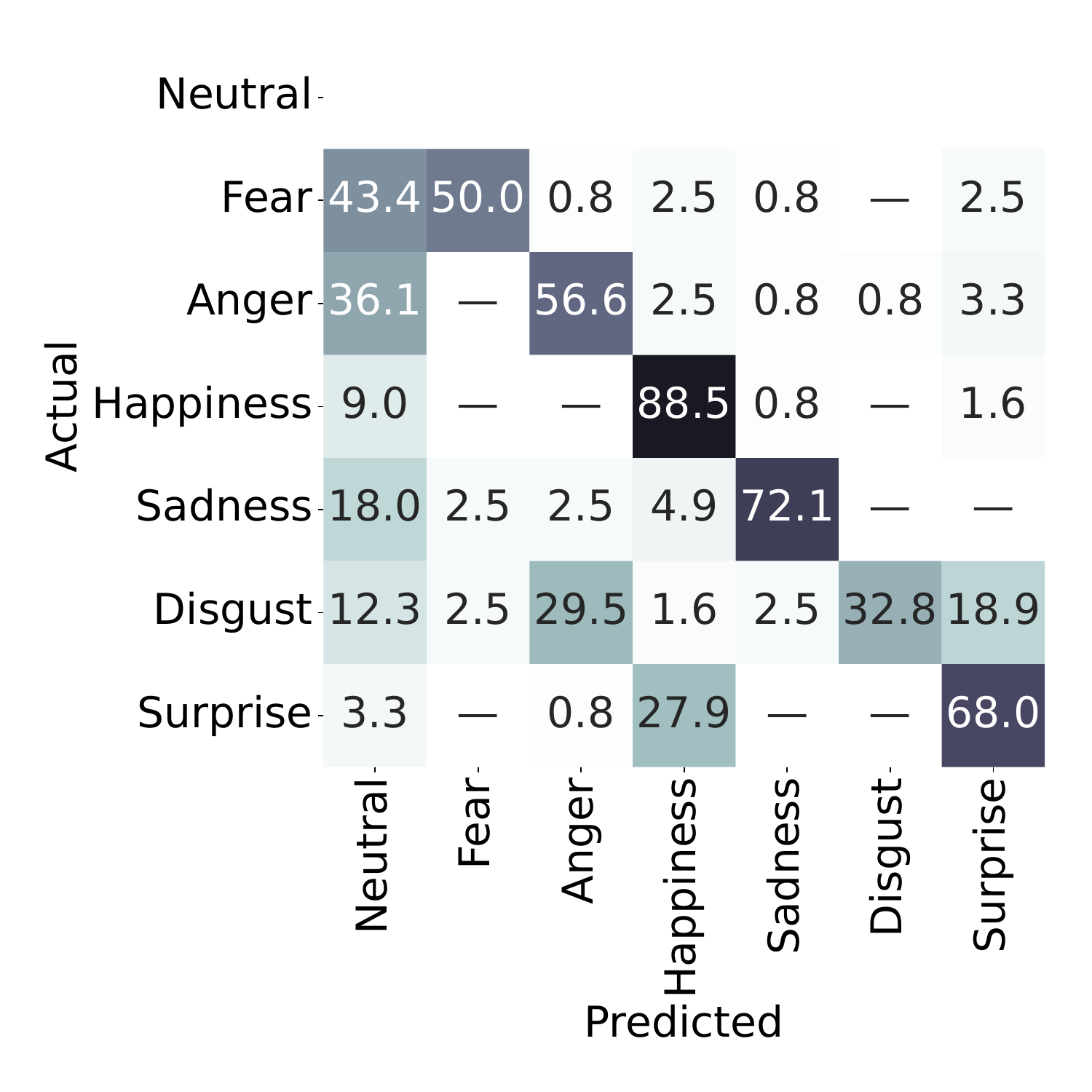}
    \caption{Mixtral}
\end{subfigure}
\hfill
\begin{subfigure}{0.24\textwidth}
    \includegraphics[width=\textwidth]{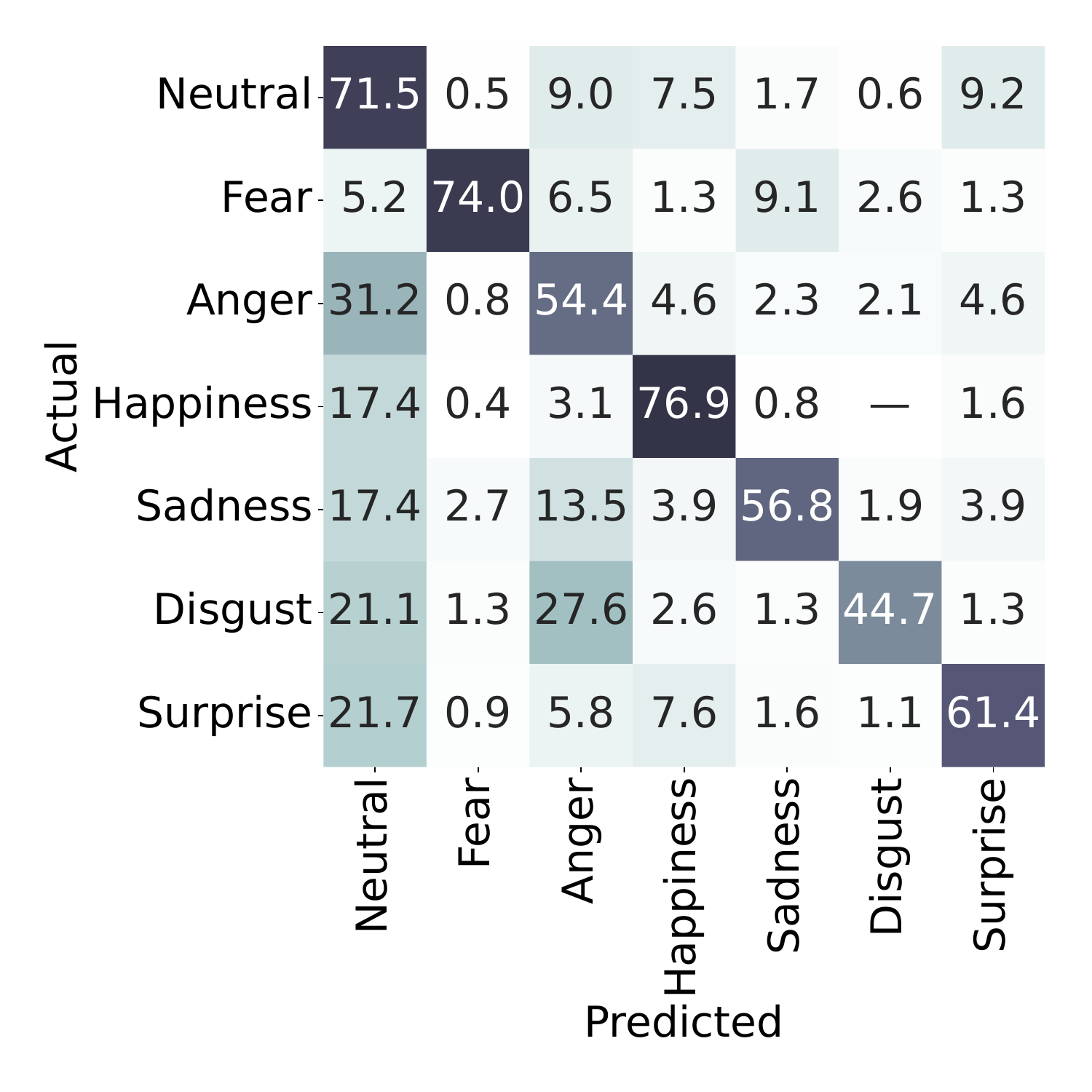}
    \caption{GoEmotions}
\end{subfigure}
\caption{Confusion matrices showing the performance (in \%) of the fine-tuned RoBERTa baseline 
on the synthesised benchmarks, 
generated 
by LLaMA2, Mistral, and Mixtral, respectively, in addition to the GoEmotions test benchmark. } 
\label{fig:confusion-roberta-generated}
\end{figure}

\Cref{fig:confusion-roberta-generated} depicts 
the confusion matrices 
obtained with the 
RoBERTa 
baseline model 
on 
the LLaMA2-, the Mistral-, and the Mixtral-generated sets. We also include its performance on the test set of GoEmotions as a reference. 
A common issue with the RoBERTa model is 
the confusion among anger and disgust. 
Analysing the confusion matrices, we also observe an interesting effect: most of the model's mispredictions are assigned to the neutral class.

\subsubsection{Analysis}
\label{sec:language_analysis}

In this section, we analyse the zero-shot sentiment analysis capabilities of the following LLMs: Mistral, Mixtral, and two versions of LLaMA2 (7 billion and 13 billion parameters). We assess their zero-shot capabilities on the test partition of the GoEmotions dataset. We design a prompt that requires the selected LLMs to predict the corresponding emotion, including the neutral state (\cf \Cref{tab:gpt-prompts}). Prompt engineering is crucial for influencing LLMs, as it enhances nuanced responses and ensures more accurate behaviour~\cite{zhou2022large}. To minimise randomness and increase confidence in the predictions, we reduce the temperature setting to 0.1. This lower temperature sharpens the probability distribution, ensuring that the predicted classes reflect those that the LLMs are predicting with the highest probabilities~\cite{amin2024prompt}. However, the LLM outputs sometimes include irrelevant or multiple emotions. To address this without intrusively altering the model's outputs, we select the first listed emotion as the most reliable prediction. This approach aligns with the operational principle of decoder-based models, where the first valid emotion is mathematically the one with the highest confidence score, thus considered the valid class prediction.

\begin{table}[!t]
    \centering
    \caption{
    Prompts 
    to use 
    LLMs for zero-shot emotion 
    recognition, following a similar pattern 
    as in \cite{amin2023wide} and \cite{amin2024prompt}.}
    \begin{tabular}{l}
    \toprule
    \textbf{Prompt template} \\
    \hline
         
    You are an expert at affective computing. 
Given a text by the user, analyze which emotion is \\most dominant in the given text. 
Only classify one of the seven Ekman emotions, namely:\\ `neutral', `fear', `anger', `happiness', `sadness', `disgust', `surprise'. You are only allowed \\to answer with EXACTLY ONE word corresponding to the aforementioned seven emotions. \\
In case of multiple emotions, use ONLY the ONE emotion you are most confident about.
\\\\
Use the following format:\\
* You are only allowed to answer with one of the following seven words: \\
``neutral'', ``fear'', ``anger'', ``happiness'', ``sadness'', ``disgust'', ``surprise''. \\
* Don't write an explanation of the answer.\\
* Don't write things like ``My guess is...", or ``I think ...". 
Just write the emotion, \\ and nothing else.\\
\bottomrule

    \end{tabular}
    \label{tab:gpt-prompts}
\end{table}

\Cref{tab:go-emotions-analysis} summarises the comparative performance of the tested LLMs. We include the performance of the RoBERTa baseline model trained on the GoEmotions dataset (\cf \Cref{subsec:language-generation}) for benchmarking purposes. The first observation from the obtained results is that the UAR scores obtained by all the investigated LLMs surpass the chance level (14.3\,\%), underscoring the emergent affective capabilities of the LLMs tested in a zero-shot manner. Nevertheless, none of the LLMs outperforms the RoBERTa baseline model, fine-tuned on the GoEmotions dataset, which confirms the advantage of model-specific tuning. Nevertheless, it is worth highlighting that the difference in the UAR scores obtained by the best-performing GPT-3.5 and GPT-4 models in comparison to the RoBERTa baseline model is around 15\,\%. This is an interesting result, as these models have not been trained on the GoEmotions dataset, but still obtained a 
competitive 
performance on the emotion recognition task, reinforcing, one more time, the emergent affective capabilities of the models. 
%
%

\begin{table*}[!t]
    \centering
    \caption{Performance scores of the different LLMs tested on a zero-shot fashion for recognising the corresponding emotion on the sentences belonging to the test partition of the GoEmotions dataset. 
    }
    \scalebox{0.76}{
\begin{tabular}{lrrrrrrrrr}
\toprule

\multirow{2}{*}{\textbf{Model}} &
\multicolumn{7}{c}{\textbf{Recall (\%)}} & \textbf{UAR} & \textbf{ACC} \\
%
\cmidrule{2-8}
& \textbf{Neutral} & \textbf{Fear} & \textbf{Anger} & \textbf{Happiness} & \textbf{Sadness} & \textbf{Disgust} & \textbf{Surprise} & \textbf{(\%)} & \textbf{(\%)} \\

\hline

LLaMA2-7B  & 4.51  & 33.77 & 42.31 & 66.60 & 39.00 & 40.79 & 58.04 & 40.72 & 38.78 \\
LLaMA2-13B & 11.89 & 40.26 & \textbf{67.31} & 69.73 & 37.45 & 40.79 & 28.12 & 42.22 & 42.39 \\
Mistral    & 57.95 & 57.14 & 47.31 & 22.08 & \textbf{58.30} & 40.79 & 9.60  & 41.88 & 39.20 \\
Mixtral    & 53.50 & 66.23 & 48.65 & 52.85 & 50.19 & 30.26 & 14.96 & 45.24 & 48.59 \\
GPT-3.5    & 23.84 & \textbf{75.32} & 56.35 & 65.54 & 40.54 & 53.95 & 18.08 & 47.66 & 43.85 \\
GPT-4      & 38.42 & 68.83 & 42.50 & 50.28 & 36.68 & \textbf{65.79} & 26.56 & 47.01 & 42.74 \\
\hline
RoBERTa    & \textbf{71.53} & 74.03 & 54.42 & \textbf{76.86} & 56.76 & 44.74 & \textbf{61.38} & \textbf{62.82} & \textbf{69.22} \\
\bottomrule
\end{tabular}
}
    
    \label{tab:go-emotions-analysis}
\end{table*}


\subsection{The Speech Modality Has (Not Yet) Changed}
\label{sec:speech}

\subsubsection{Generation}
Research on generating affective speech has been conducted for more than three decades~\cite{triantafyllopoulos2023overview}. 
The first 
approaches were rule-based, while contemporary methods typically rely on deep learning. A detailed overview can be found in \cite{triantafyllopoulos2023overview}. All of these methods are explicitly engineered to produce emotional speech. Thus, this line of research can be dubbed as a subfield of `traditional' Affective Computing, while technically belonging to the Text-To-Speech (TTS) field. In recent years, similar to the developments in Natural Language Processing (NLP) and Computer Vision (CV), research on TTS 
has heavily 
been 
influenced by the success of the Foundation Model (FM) paradigm~\cite{yang2023uniaudio, leng2023prompttts}. 

UniAudio~\cite{yang2023uniaudio} is a Transformer-based general-purpose audio synthesis system. It is pretrained on 
seven 
generative audio tasks, including TTS. In its pretrained state, however, it does not support affective speech synthesis. The authors demonstrate that their pretrained model serves as a basis for adaptation to different downstream tasks which opens up the possibility to equip the model with affective speech synthesis capabilities by mere finetuning. Another recent generative audio 
FM, PromptTTS2~\cite{leng2023prompttts}, synthesises speech based on text prompts that include descriptions of the voice to be generated. The controllable attributes of the synthesised speech must be defined during the training time. The authors of \cite{leng2023prompttts} do not investigate emotion as one such attribute, though their proposed framework would permit this.
Models such as UniAudio and PromptTTS2 can thus be 
categorised 
as generative audio 
FMs 
that could be adapted to emotional speech synthesis with moderate effort. To the best of our knowledge, no evidence for affective speech synthesis as an emergent capability of such models has been published so far. However, the demos for GPT-4o\footnote{\href{https://openai.com/index/hello-gpt-4o/}{https://openai.com/index/hello-gpt-4o/}} claim affective speech synthesis capabilities. At the moment, though, neither a technical report on GPT-4o, nor a systematic evaluation of affective speech produced by GPT-4o is available.  
Considering the development towards releasing large pretrained models in the NLP and CV areas, we assume that in the near future powerful speech synthesis models will also be made publicly available and, hence, investigated more thoroughly, 
allowing us to carry out according experiments to the above for vision and linguistics.
In addition, we expect a continuing trend toward truly multimodal 
FMs 
that may not just take in but also produce natural speech with controllable prosodic properties. Several multimodal models that also produce audio output data have been proposed, for an overview see \cite{wu2023multimodal, zhang2024mm}. To the best of our knowledge, none of them exhibit emotional speech synthesis capabilities. We expect that emergent affective speech synthesis will eventually be achieved via a multimodal approach. As shown in the previous sections, large pretrained vision and language models already encode affective information. Multimodal approaches leveraging such pretrained models in combination with multimodal affect-related data may learn to associate affective speech characteristics with corresponding affective states as expressed in the visual and the textual data.

\subsubsection{Analysis}
A system capable of analysing arbitrary affective properties of speech data without any tuning must ingest both audio and text inputs. Several 
FM 
approaches fulfilling this requirement have been proposed. However, the vast majority of them are not pretrained on speech data at all. 

Examples include AnyMAL~\cite{moon2023anymal}, X-InstructBLIP~\cite{panagopoulou2023x}, and ModaVerse~\cite{wang2024modaverse}.
In only a few models, speech is part of the pretraining data. QWEN-Audio's training data comprises several labelled speech datasets, including emotionality already. Hence, QWEN-Audio~\cite{chu2023qwen} in the proposed form is not a candidate for exploring `emergent' 
affective 
recognition capabilities. X-LLM~\cite{chen2023x} processes video, text, and audio 
inputs and is explicitly designed to process speech. The authors, however, do not report any experiments related to predicting affect in speech. As of now, the pretrained X-LLM model is not publicly available, 
hence, unfortunately again, not allowing us to carry out experiments analogous to the vision and the linguistic ones.

Similar to the affective speech synthesis problem, near-future multimodal 
FMs 
can be expected to be capable of analysing affective speech in a zero-shot manner, even if not explicitly pretrained in this regard. As of now, however, we are not aware of any such model.  

\subsection{The Evaluation Is Changing}

One of the reasons to understand the impressive performance of currently available Foundation Models (FM) is that they use massive amounts of data from ``the Internet'' for training. Nevertheless, the indiscriminate use of data poses the following challenge to the scientific community: can we guarantee that the data we feed to these models for testing -- or even for evaluation -- has not been used for training? 
In case of a negative answer, how fair and representative of the model capabilities can standard evaluation metrics be? Although we do not have a concrete answer yet, we hope these challenges engage the research community into looking for methods and metrics that allow a proper scientific evaluation of these emerging 
FMs 
in the field of Affective Computing. 
As is, the current state of such models in Affective Computing may resemble a shell game: many different tools and approaches are shuffled and mixed until some partially less, partially more convincing performances are obtained. Especially because it is the popular `Big Six' Ekman emotions we considered herein, chances are high that the models only reverberate with what they have 
already seen. 
Testing on more subtle models such as the dimensional approach or less considered affective states is therefore urgently needed.

\section{Concerns and Regulations Have Changed}
\label{sec:concerns}

In April 2021, the European Commission presented the AI Act: the first-ever legal framework worldwide to regulate the use of AI-based technologies in the European territory. The Act was endorsed by all Member States in February 2024 after being approved by the EU Council. The regulation follows a risk-based approach, so instead of regulating specific systems and applications, it defines measures and requirements based on a classification system that encapsulates varying degrees of risks posed by the AI systems. The proposed classification system spans 
four different levels of risk: unacceptable, high, limited, and minimal. 
According to the final version of the Act\footnote{\url{https://www.euaiact.com/}}, systems that fall into the unacceptable risk category will be prohibited. 

This regulation is of special interest for the Affective Computing community, since it singles out 
affective 
systems in several ways. The regulation defines an emotion recognition system in its Article 3 (34) as an ``AI system for the purpose of identifying or inferring emotions or intentions of natural persons on the basis of their biometric data''. Thus, any emotion recognition system using speech or facial data, as well as other physical signals such as 
the electroencephalogram, falls under this definition. Article 5 enumerates different practices and systems considered as prohibited, including the placing on the market or the use of AI systems to infer emotions of a natural person in the 
workplace 
or in 
education institutions. Whilst this point only prohibits emotion recognition systems in two specific contexts (workplace and education), the list of high-risk systems presented in Annex III includes ``AI systems intended to be used for emotion recognition'' in the category of biometric systems. Article 6 provides a nuance on the referred systems, 
stating that AI systems ``shall not be considered as high risk if they do not pose a significant risk of harm to the health, safety, or fundamental rights of natural persons''. However, this exception shall not apply for systems performing profiling as defined in Article 4 of the General Data Protection Regulation (GDPR). Considering the definition of profiling, it is unclear whether any emotion recognition system, even in non-harmful applications such as entertainment, could be considered of limited risk, except for those based on non-identifiable data. In any case, Article 52 imposes that any system including an emotion recognition component must notify the users of the operation of such system, notwithstanding its risk.

In this context, the regulation imposes several obligations that high-risk systems shall comply with, and for which the provider (\ie, a natural person/agency that develops and places on the market the AI system) is responsible. 
These obligations are detailed across Chapters 2-5 of Title III -- more than 40 articles -- and include conducting post-market surveillance of risk, informing the competent authorities about the product, or providing technical documentation 
of both the system and the data used in development (\ie, data governance), among others. Note that the definition of provider is quite vague from a research perspective, so it is unclear how this will affect research on Affective Computing. Article 2 (5a) specifies that the regulation does not apply to models developed for the sole purpose of scientific research, but by the time the model is made available, researchers may face some of the previously cited obligations depending on how it is used, after which they may become providers. By considering cases in which the system is made available free of charge, the regulation seems to cover open-source systems and situations in which the source code is made publicly available. However, this does not seem to be the case when two academic institutions share code in a confidential manner for academic purposes. 

Attending now to Foundation Models (FM), which are referred in the text as ``General-purpose AI models'', Article 52 requires any AI-synthesised content to be marked as such, which clearly covers 
generation 
of emotional samples with models such as LLaMA, or SD, 
among others. 
Besides, Article 52 includes several subparagraphs with the obligations that providers shall meet when deploying this kind of systems, such as providing documentation on how it was trained and validated, and the content used for that purpose. In addition, 
an FM 
may be considered to pose systemic risk if it has high computational capabilities, or if it is marked as such by the Commission. In this case, providers shall comply with more obligations, including testing and mitigating foreseeable risks, data governance measures, maintaining appropriate levels of performance and interpretability. In this context, there are some concerns that current 
FMs 
do not comply with all the measures required~\cite{bommasani2023eu}, 
hence accepting to slow down innovation and AI development (particularly) in Europe, to assure highest safety and ethical standards.



Beyond, emotion recognition based on physiological data has been less investigated than other modalities, such as 
facial expressions or speech. 
Some types of physiological signals include electroencephalogram (EEG), electrocardiogram (ECG), electromyogram (EMG), electrodermal activity (EDA) or galvanic skin response (GSR), respiration rate (RSP), and pulse rate. 
Emotions are complex and sometimes cannot be solely recognised by analysing speech or image data. 
It is quite easy for people to control -- or even hide -- their real current emotional state under certain circumstances, mainly 
because of social pressure. For instance, people may pretend to smile and laugh while they are feeling sad or angry~\cite{shu2018review}.
%
Ethical considerations regarding privacy and data security are of high importance when analysing physiological data for 
affective 
purposes to avoid both: `(Full) Affective Mind Reading' and `Affective Mind Writing'.

Many further challenges and ethical concerns remain and will also become more apparent once 
such 
technology is broadly used -- hopefully, the community can provide technical and legal means of protection to 
ensure 
we are all enjoying the positive side of Affective Computing only.

%

\section{Outlook and Conclusions}
\label{sec:conclusions}

In this paper, we analysed the affective capabilities of currently available Foundation Models (FM) exploiting the vision, the linguistics, and the speech (acoustic) modalities. While the affective generation 
and analysis 
capabilities of the vision- and the linguistic-based FMs are plausible, the affective generation and analysis of speech-based FMs is not yet mature enough. Nonetheless, it is reasonable to imagine a not-too-distant future where this technology achieves similar results as with the other two modalities. Despite not being currently available, we also envision physiological-based FMs to be developed and explored in the near future. 

One of the main outcomes of this work is the collection of two synthetically-generated affective corpora generated with FMs -- one containing facial images, the other textual sentences 
that will be publicly available. 
The models training and the analyses reported herein were performed assuming that the synthetically generated instances conveyed the prompted emotions. Nonetheless, we acknowledge this could not always be the case. To overcome this limitation, we plan to run a data collection with human annotators to annotate the generated samples, assessing the affective capabilities of the selected FMs from a human perspective. 


\section*{Funding}

This project has received funding from the DFG's Reinhart Koselleck project No.\,442218748 (AUDI0NOMOUS).

\bibliography{sn-article}

\end{document}